\documentclass[a4paper]{llncs}
\usepackage{graphicx}
\usepackage{amsmath,amssymb} 
\usepackage{color}
\usepackage[width=122mm,left=12mm,paperwidth=146mm,height=193mm,top=12mm,paperheight=217mm]{geometry}

\usepackage{bm}
\usepackage{nth}
\usepackage{array}
\usepackage{lscape}

\begin{document}
  \pagestyle{headings}
  \mainmatter
  \def\ECCV18SubNumber{}  

  \title{Faster Gaze Prediction With Dense Networks and Fisher Pruning}



  \author{
    Lucas Theis \and Iryna Korshunova \and Alykhan Tejani \and Ferenc Husz\'{a}r \\
    \texttt{\{ltheis,ikorshunova,atejani,fhuszar\}@twitter.com}}
  \institute{Twitter}

  \maketitle

  \begin{abstract}
    Predicting human fixations from images has recently seen large improvements by leveraging deep
    representations which were pretrained for object recognition. However, as we show in this paper,
    these networks are highly overparameterized for the task of fixation prediction. We first
    present a simple yet principled greedy pruning method which we call Fisher pruning. Through a
    combination of knowledge distillation and Fisher pruning, we obtain much more runtime-efficient
    architectures for saliency prediction, achieving a \textnormal{10x} speedup for the same AUC performance as a
    state of the art network on the CAT2000 dataset. Speeding up single-image gaze prediction is
    important for many real-world applications, but it is also a crucial step in the development of
    video saliency models, where the amount of data to be processed is substantially larger.
  \end{abstract}

  \section{Introduction}

  The ability to predict the gaze of humans has many applications in computer vision and related
  fields. It has been used for image cropping \cite{ardizzone2013cropping}, improving video compression \cite{feng2012encoding}, and as a tool
  to optimize user interfaces \cite{xu2016gui}, for instance. In psychology, gaze prediction models are used to shed light on
  how the brain might process sensory information \cite{koch1985neuro}. Recently, due to advances in deep learning, human gaze prediction has
  received large performance gains. In particular, reusing image representations trained for the task of object
  recognition have proven to be very useful \cite{kuemmerer2014deepgaze,kummerer2016deepgaze2}. However, these networks are relatively slow to evaluate
  while many real-world applications require highly efficient predictions. For example, popular websites
  often deal with large amounts of images which need to be processed in a short amount of time using
  only CPUs. Similarly, improving video encoding with gaze prediction maps requires the processing of large
  volumes of data in near real-time.

  In this paper we explore the trade-off between computational complexity and gaze prediction performance.
  Our contributions are two-fold: First, using a combination of knowledge distillation \cite{hinton2015distilling} and
  pruning, we show that good performance can be achieved with a much faster
  architecture, achieving a 10x speedup for the same generalization performance in terms of AUC. Secondly, we provide a principled derivation for the
  pruning method of Molchanov et al. \cite{molchanov2017pruning}, extend it, and show that our extension works well when applied to gaze
  prediction. We further discuss how to choose the trade-off between performance and
  computational cost and suggest methods for automatically tuning a weighted combination of the
  corresponding losses, reducing the need to run expensive hyperparameter searches.

  \section{Fast Gaze Prediction Models}
    Our models build on the recent state-of-the-art model \textit{DeepGaze~II} \cite{kummerer2016deepgaze2}, which we first review before discussing our approach to speeding it up. The backbone of DeepGaze~II is formed by VGG-19~\cite{simonyan2014vgg},
    a deep neural network pre-trained for object recognition. Feature maps are extracted from
    several of the top layers, upsampled, and concatenated. A \textit{readout network}
    with $1 \times 1$ convolutions and ReLU nonlinearities \cite{nair2010relu} takes in
    these feature maps and produces a single output channel, implementing a point-wise nonlinearity.
    This output is then blurred with a Gaussian filter, $G_\sigma$, followed by the addition of a center bias to take into account the tendencies of observers to fixate on pixels near the image center. This center bias
    is computed as the marginal log-probability of a fixation landing on a given pixel, $\log Q(x, y)$,
    and is dataset dependent. Finally, a softmax operation is applied to produce a normalized
    probability distribution over fixation locations, or \textit{saliency map}:
    \begin{align}
      Q(x, y \mid \mathbf{I}) \propto \exp\left( R(U(F(\mathbf{I}))) * G_\sigma + \log Q(x, y) \right)
    \end{align}
    Here, $\mathbf{I}$ is the input image, $F$ extracts feature maps, $U$ bilinearly upsamples the feature
    maps and $R$ is the readout network.

    To improve efficiency, we made some minor modifications in our reimplementation of DeepGaze~II.
    We first applied the readout network and then bilinearly upsampled the one-dimensional output of the
    readout network, instead of upsampling the high-dimensional feature maps. We also used separable filters
    for the Gaussian blur. To make sure the size of the saliency map matches the size of the input image,
    we upsample and crop the output before applying the softmax operation.

    We use two basic alternative architectures providing different trade-offs between computational
    efficiency and performance. First, instead of VGG-19, we use the faster VGG-11 architecture~\cite{simonyan2014vgg}. As
    we will see, the performance lost by using a smaller network can for the most part be compensated by fine-tuning the
    feature map representations instead of using fixed pre-trained representations. Second, we try
    DenseNet-121~\cite{huang2017densely} as a feature extractor. DenseNets have been shown to be more efficient, both computationally and in terms of parameter efficiency, when compared to state-of-the-art networks in the
    object recognition task~\cite{huang2017densely}.

    Even when starting from these more parameter efficient pre-trained models, the resulting gaze
    prediction networks remain highly over-parametrized for the task at hand. To further decrease
    the number of parameters we turn to pruning: greedy removal of redundant parameters or feature
    maps. In the following section we derive a simple, yet principled, method for greedy network
    pruning which we call \textit{Fisher pruning}.

  \subsection{Fisher Pruning}
    \label{sec:fisher_pruning}
    Our goal is to remove feature maps or parameters which contribute little to the overall performance of the
    model. In this section, we consider the general case of a network with parameters $\bm{\theta}$ trained to minimize a cross-entropy loss,
    \begin{align}
      \mathcal{L}(\bm{\theta}) = \mathbb{E}_P\left[ -\log Q_{\bm{\theta}}(\mathbf{z} \mid \mathbf{I}) \right],
    \end{align}
    where $\mathbf{I}$ are inputs, $\mathbf{z}$ are outputs, and the expectation is taken with
    respect to some data distribution~$P$. We first consider pruning single parameters $\theta_k$.
    For any change in parameters $\mathbf{d}$, we can approximate the corresponding change in loss
    with a \nth{2} order approximation around the current parameter value $\bm{\theta}$:
    \begin{align}
      \mathbf{g} = \nabla \mathcal{L}(\bm{\theta}), \quad & \quad
      \mathbf{H} = \nabla^2 \mathcal{L}(\bm{\theta}), \\
      \mathcal{L}(\bm{\theta} + \mathbf{d}) - \mathcal{L}(\bm{\theta}) &\approx
      \mathbf{g}^\top \mathbf{d} + \frac{1}{2} \mathbf{d}^\top \mathbf{H} \mathbf{d}
    \end{align}
    Following this approximation, dropping the $k$th parameter (setting $\theta_k=0$) would lead to the following increase
    in loss:
    \begin{align}
      \mathcal{L}(\bm{\theta} - \theta_k \mathbf{e}_k) - \mathcal{L}(\bm{\theta}) \approx
      -g_k \theta_k + \frac{1}{2} H_{kk} \theta_k^2,
      \label{eq:approximation}
    \end{align}
    where $\mathbf{e}_k$ is the unit vector which is zero everywhere except at its $k$th entry,
    where it is 1. Following related methods which also start from a \nth{2} order approximation
    \cite{lecun1990optimal,hassibi1993second},
    we assume that the current set of parameters is at a local optimum and that the \nth{1} term
    vanishes as we average over a dataset of input images. In practice, we found that including the first term actually reduced the performance of the pruning method. For the diagonal of the Hessian, we use the approximation
    \begin{align}
      H_{kk} \approx \mathbb{E}_P\left[ \left( \frac{\partial}{\partial \theta_k} \log Q_{\bm{\theta}}(\mathbf{z} \mid \mathbf{I}) \right)^2 \right],
      \label{eq:hessian}
    \end{align}
    which assumes that $Q_{\bm{\theta}}(\mathbf{z} \mid \mathbf{I})$ is close to $P(\mathbf{z} \mid \mathbf{I})$
    (see Supplementary Section~1 for a derivation).
    Eqn.\,\eqref{eq:hessian} can be viewed as an empirical estimate of the Fisher information of $\theta_k$,
    where an expectation over the model is replaced with real data samples.
    If $Q$ and $P$ are in fact equal and the model is twice differentiable with respect to
    parameters $\theta$, the Hessian reduces to the Fisher information matrix and the
    approximation becomes exact.

    If we use $N$ data points to estimate the Fisher information, our approximation of the increase in loss
    becomes
    \begin{align}
      \Delta_k = \frac{1}{2N} \theta_k^2 \sum_{n = 1}^N g_{nk}^2,
      \label{eq:pruning_signal}
    \end{align}
    where $\mathbf{g}_n$ is the gradient of the parameters with respect to the $n$th data point.
    In what follows, we are going to use this as a pruning signal to greedily remove parameters one-by-one where this estimated increase in loss is smallest.

    For convolutional architectures, it makes sense to try to prune entire feature maps instead of
    individual parameters, since typical implementations of convolutions may not be able to exploit
    sparse kernels for speedups. Let $a_{nkij}$ be the activation of the $k$th feature map at spatial location $i,j$ for the $n$th datapoint.
    Let us also introduce a binary mask $\mathbf{m}\in\{0,1\}^K$ into the network which modifies the activations $a_{nkij}$ of each feature map $k$ as follows:
    \begin{align}
      \tilde{a}_{nkij} = m_k a_{nkij}.
      \label{eqn:masked_activations}
    \end{align}

    The gradient of the loss for the $n$th datapoint with respect to $m_k$ is
    \begin{align}
      g_{nk} &= -\sum_{ij} a_{nkij} \frac{\partial}{\partial a_{nkij}} \log Q(\mathbf{z}_n \mid \mathbf{I}_n)
      \label{eqn:gnk_colvolutional}
    \end{align}
    and the pruning signal is therefore $\Delta_k = \frac{1}{2N}\sum_n g_{nk}^2$, since $m_k^2 = 1$ before
    pruning. The gradient with respect to the activations is available during the
    backward pass of computing the network's gradient and the pruning signal can therefore be
    computed at little extra computational cost.

    We note that this pruning signal is very similar to the one used by Molchanov et al. \cite{molchanov2017pruning} -- which uses
    absolute gradients instead of squared gradients and a certain normalization of the pruning signal -- but our derivation provides a more principled
    motivation. An alternative derivation which does not require $P$ and $Q$ to be close is
    provided in Supplementary Section~2.

  \subsection{Regularizing Computational Complexity}
    In the previous section, we have discussed how to reduce the number of parameters or feature
    maps of a neural network. However, we are often more interested in reducing a network's computational
    complexity. That is, we are trying to solve an optimization problem of the form
    \begin{align}
      \underset{\bm{\theta}}{\text{minimize}} \quad \mathcal{L}(\bm{\theta}) \quad \text{subject to} \quad \mathcal{C}(\bm{\theta}) < K,
    \end{align}
    where $\bm{\theta}$ here may contain the weights of a neural network but may also contain a binary
    mask describing its architecture. $\mathcal{C}$ measures the computational complexity of the network. During
    optimization, we quantify the computational complexity in terms of floating point operations.
    For example, the number of floating point operations of a convolution with a bias term, $K \times K$ filters,
    $C_\text{in}$ input channels, $C_\text{out}$ output channels, and producing a feature map with spatial extent $H \times W$ is
    given by
    \begin{align}
      H \cdot W \cdot C_\text{out} \cdot (2 \cdot C_\text{in} \cdot K^2 + 1).
      \label{eq:comp_cost}
    \end{align}
    Since $H$ and $W$ represent the size of the output, this formula automatically takes into
    account any padding as well as the stride of a convolution. The total cost of a network is the
    sum of the cost of each of its layers.

    To solve the above optimization problem, we try to minimize the Lagrangian
    \begin{align}
      \mathcal{L}(\bm{\theta}) + \beta \cdot \mathcal{C}(\bm{\theta}),
      \label{eq:combined_cost}
    \end{align}
    where $\beta$ controls the trade-off between computational complexity and a model's performance.
    We compute the cost of removing a parameter or feature map as
    \begin{align}
      \mathcal{L}(\bm{\theta} - \theta_k \mathbf{e}_i) - \mathcal{L}(\bm{\theta}) + \beta \cdot \left(
      \mathcal{C}(\bm{\theta} - \theta_k \mathbf{e}_i) - \mathcal{C}(\bm{\theta}) \right),
      \label{eq:feature_cost}
    \end{align}
    where the increase in loss is estimated as in the previous section. During training, we
    periodically estimate the cost of all feature maps and greedily prune feature maps which
    minimize the combined cost. When pruning a feature map, we expect the loss to go up but the
    computational cost to go down. For different $\beta$, different architectures will become
    optimal solutions of the optimization problem.

  \subsection{Automatically Tuning $\beta$}
    How should $\beta$ be chosen? One option is to treat it like any hyperparameter and
    to train many models with different values of $\beta$. In some settings, this may not be
    feasible. In this section, we therefore discuss an approach which allows generating many models of
    different complexity in a single training run.

    For a given $\beta$, a feature should be pruned if Equation~\ref{eq:feature_cost} is negative, that is, when doing so
    reduces the overall cost because it decreases the computational cost more than it increases the cross-entropy:
    \begin{align}
      \Delta\mathcal{L}_i + \beta \cdot \Delta \mathcal{C}_i \leq 0
      \label{eq:criterion}
    \end{align}
    We propose choosing the smallest $\beta$ such that after removing all features with negative
    or zero pruning signal, a reduction in either a desired number of features or total
    computational cost is achieved.

    The threshold for pruning feature map $i$ is reached when setting the weight to
    \begin{align}
      \beta_i = -\frac{ \Delta\mathcal{L}_i }{ \Delta \mathcal{C}_i }.
      \label{eq:auto_weight}
    \end{align}
    Consider pruning only a single feature map. The smallest $\beta$ such that
    Equation~\ref{eq:criterion} is satisfied for at least one feature map is given by $\beta^* =
    \min_i \beta_i$. For $i$ with $\beta_i \neq \beta^*$, we have
    \begin{align}
      \Delta\mathcal{L}_i + \beta^* \cdot \Delta \mathcal{C}_i
      &= (-\beta_i + \beta^*) \Delta \mathcal{C}_i > 0,
    \end{align}
    since $\Delta \mathcal{C}_i < 0$. That is, these feature maps should not be pruned, which means that $\beta^*$ is a reasonable
    choice if we only want to prune 1 feature map. We propose a greedy strategy, where in each
    iteration of pruning only 1 feature map is targeted and $\beta^*$ is used as a weight. Note that
    we can equivalently use the $\beta_i$ directly as a hyperparameter-free pruning signal. This signal is intuitive,
    as it picks the feature map whose increase in loss is small relative to the decrease in computational cost.

    Another possibility is to automatically tune $\beta$ such that the total reduction in cost reaches
    a target if one were to remove all feature maps with negative pruning signal
    (Equation~\ref{eq:criterion}). However, we do not explore this option further in this paper.

  \subsection{Training}
    \begin{figure}
      \centering
      \includegraphics[width=10cm]{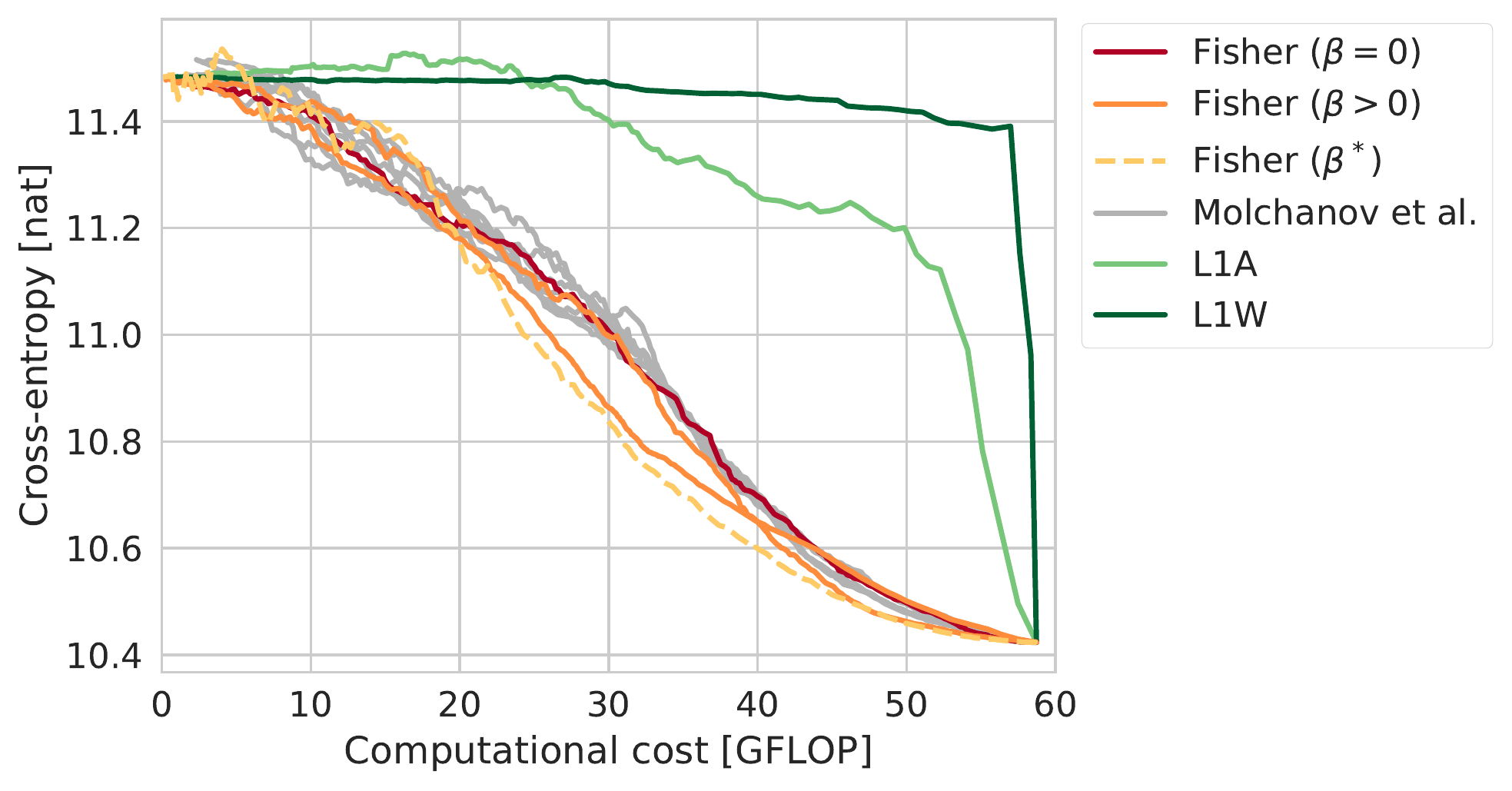}
      \caption{Comparison of different pruning methods applied to FastGaze without retraining. We compared Fisher pruning
      to two baselines and the method of Molchanov et al.~\cite{molchanov2017pruning}. For the
      latter, we ran the normalized/unnormalized and regularized/unregularized variants (gray lines).
      This plot shows that naive methods do not work well (green lines), that using different
      regularization weights is important, and that updating estimates of the computational cost during pruning
      is important for optimal performance (orange lines).}
      \label{fig:pruning_comparison}
    \end{figure}

    Our saliency models were trained in several steps. First, we trained a DeepGaze~II model using Adam~\cite{kingma2015adam} with
    a batch size of 8 and an initial learning rate of 0.001 which was slowly decreased over the course of training.
    As in~\cite{kummerer2016deepgaze2}, the model was first pre-trained using the SALICON dataset~\cite{jiang2015salicon} while using the MIT1003 dataset~\cite{judd2009mit} for
    validation. The validation data was evaluated every 100 steps and training was stopped when
    the cross-entropy on the validation data did not decrease 20 times in a row. The parameters with
    the best validation score observed until then were saved. Afterwards, the MIT1003 dataset
    was split into 10 training and validation sets and used to train 10 DeepGaze~II models again
    with early stopping.

    The ensemble of DeepGaze~II models was used to generate an average saliency map
    for each image in the SALICON dataset. These saliency maps were then used for knowledge
    distillation~\cite{hinton2015distilling}. This additional data allows us to not only train the readout network of our own
    models, but also fine-tune the underlying feature representation. We used a weighted combination
    of the cross-entropy for MIT1003 and the cross-entropy with respect to the DeepGaze~II saliency
    maps, using weights of 0.1 and 0.9, respectively.

    After training our models to convergence, we start pruning the network. We accumulated pruning
    signals (Equation~\ref{eq:pruning_signal}) for 10 training steps while continuing to update the
    parameters before pruning a single feature map. The feature map was selected to maximize the
    reduction in the combined cost (Equation~\ref{eq:feature_cost}). We tried to apply early
    stopping to the combined cost to automatically determine an appropriate number of feature maps to prune,
    however, we found that early stopping terminated too early and we therefore opted to treat
    the number of pruned features as another hyperparameter which we optimized via random search. During
    the pruning phase we used SGD with a fixed learning rate of 0.0025 and momentum of 0.9,
    as we found that this led to slightly better results than using Adam. This may be explained by a
    regularizing effect of SGD~\cite{zhang2017rethinking}.

  \subsection{Related Work}
    Many recent papers have used pretrained neural networks as feature extractors for the prediction
    of fixations~\cite{kuemmerer2014deepgaze,kruthiventi2017deepfix,kummerer2016deepgaze2,tavakoli2017,liu2016}.
    Most closely related to our work is the \textit{DeepGaze} approach \cite{kuemmerer2014deepgaze,kummerer2016deepgaze2}.
    In contrast to \textit{DeepGaze}, here we also finetune the feature representations, which
    despite the limited amount of available fixation data is possible
    because we use a combination of knowledge distillation and pruning to regularize our networks.
    Kruthiventi et al.~\cite{kruthiventi2017deepfix} also tried to finetune a pretrained network by
    using a smaller learning rate for pretrained parameters than for other parameters.
    Vig et al.~\cite{vig2014edn} trained a smaller network end-to-end but did not start from a pretrained representation and
    therefore did not achieve the performance of current state-of-the-art architectures. Similarly,
    Pan et al.~\cite{pan2016salnet} trained networks end-to-end while initializing only a few layers with
    parameters obtained from pretrained networks but have since been outperformed by DeepGaze~II
    and other recent approaches.

    To our knowledge, none of this prior work has addressed the question of how much computation
    is actually needed to achieve good performance.

    Many heuristics have been developed for pruning \cite{li2017pruning,han2015pruning,molchanov2017pruning}. More closely related to ours are
    methods which try to directly estimate the effect on the loss. \textit{Optimal brain
    damage}~\cite{lecun1990optimal}, for example, starts from a 2nd order approximation of a squared
    error loss and computes second derivatives analytically by performing an additional backward
    pass. \textit{Optimal brain surgeon}~\cite{hassibi1993second} extends this method and
    automatically tries to correct parameters after pruning by computing the full Hessian. In
    contrast, our pruning signal only requires gradient information which is already computed during
    the backward pass. This makes the proposed Fisher pruning not only more efficient but also easier to implement.

    Most closely related to our pruning method is the approach of~Molchanov et al.~\cite{molchanov2017pruning}.
    By combining a 1st order approximation with heuristics, they arrive at a very similar estimate of
    the change in loss due to pruning. We found that in practice, both estimates performed similarly
    when used without regularization (Figure~\ref{fig:pruning_comparison}). The main contribution in
    Section~\ref{sec:fisher_pruning} is a new derivation which provides a more principled motivation
    for the pruning signal.

    Unlike most papers on pruning, Molchanov et al.~\cite{molchanov2017pruning} also explicitly regularized
    the computational complexity of the network. However, their approach to regularization differs
    from ours in two ways. First, a fixed weight was used for the computational cost while pruning a different number
    of feature maps. In contrast, here we recognize that each setting of $\beta$ creates a separate
    optimization problem with its own optimal architecture. In practice, we find that the speed and
    architecture of a network is heavily influenced by the choice of $\beta$ even when pruning the same number of feature maps,
    suggesting that using different weights is important. Molchanov et al.~\cite{molchanov2017pruning}
    further only estimated the computational cost of each feature map once before starting to prune.
    This leads to suboptimal pruning, as the computational cost of a feature map changes when
    neighboring layers are pruned (Figure~\ref{fig:pruning_comparison}).

  \section{Experiments}
    \begin{figure*}[bt]
      \centering
      \includegraphics[width=9cm]{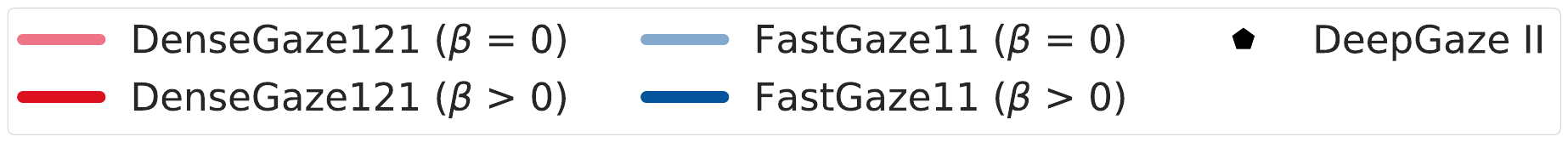}
      \includegraphics[width=6cm]{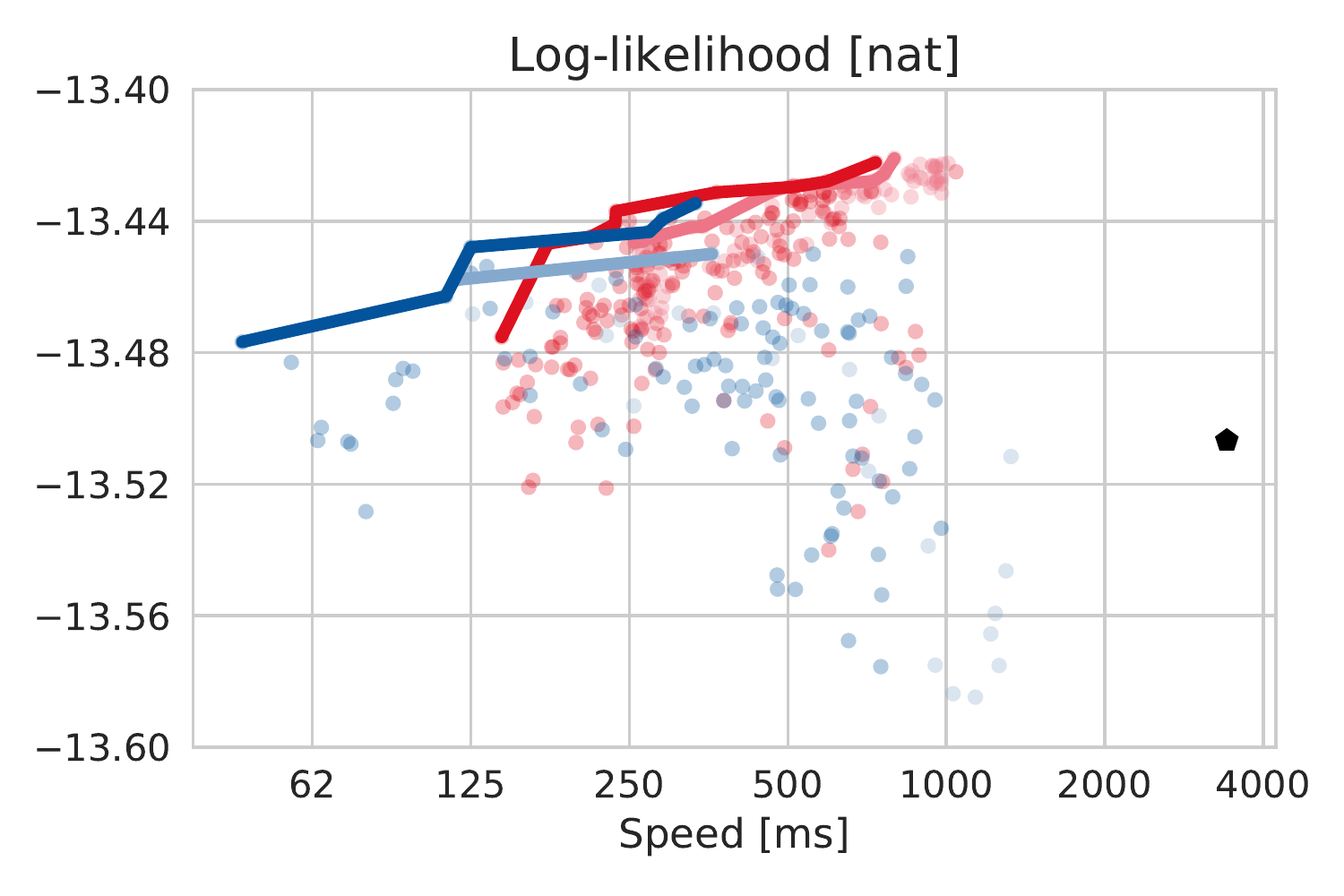}
      \includegraphics[width=6cm]{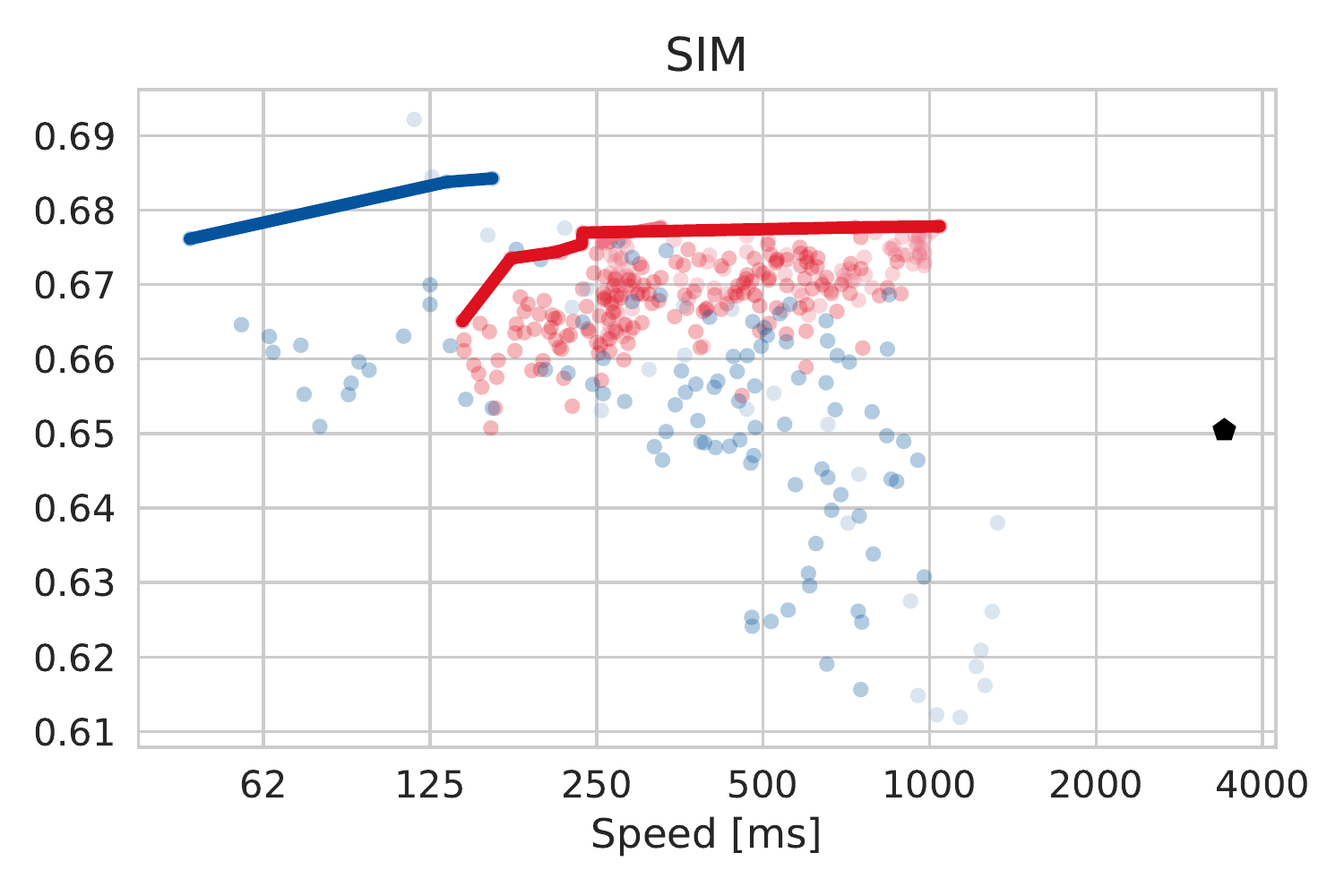}
      \includegraphics[width=6cm]{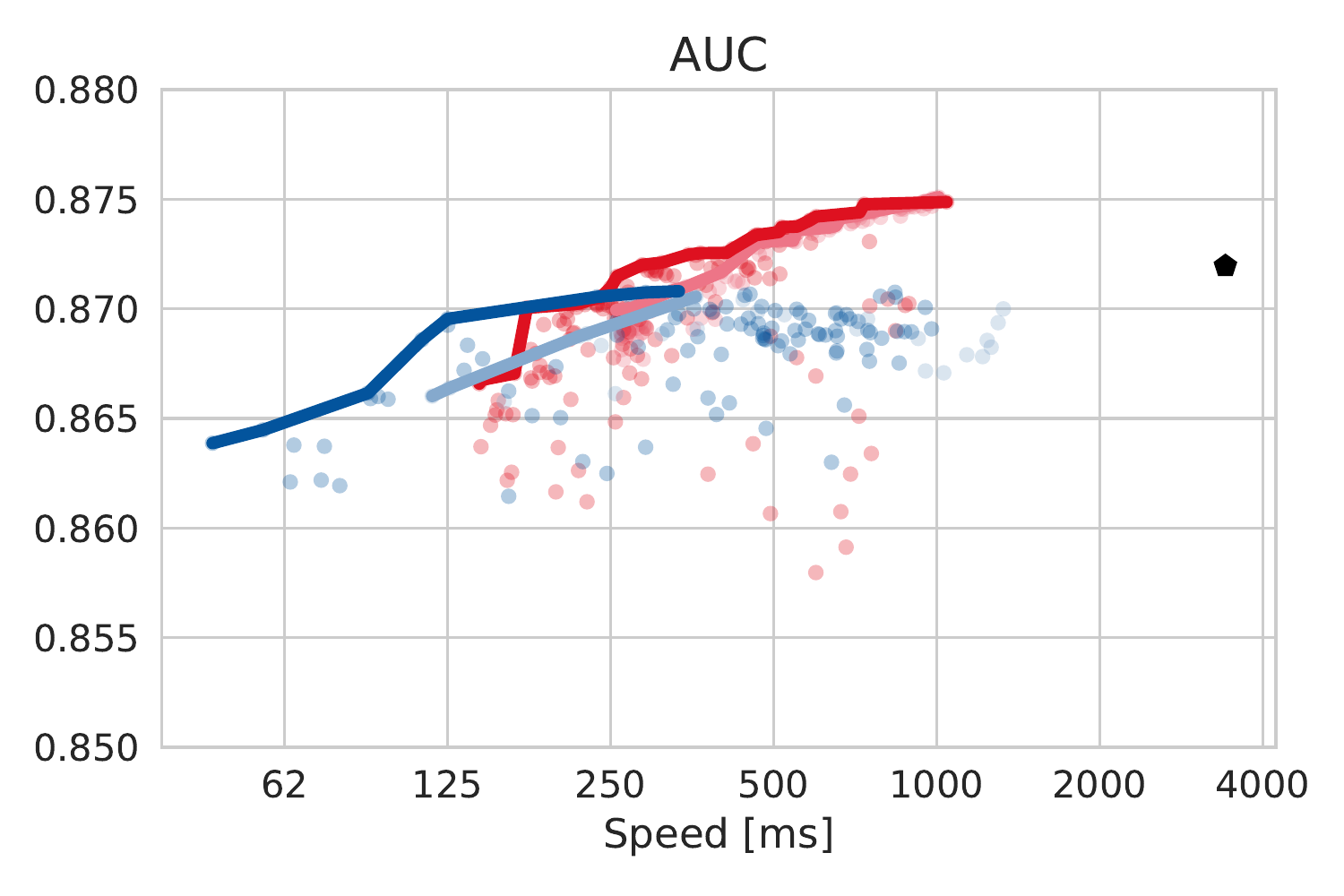}
      \includegraphics[width=6cm]{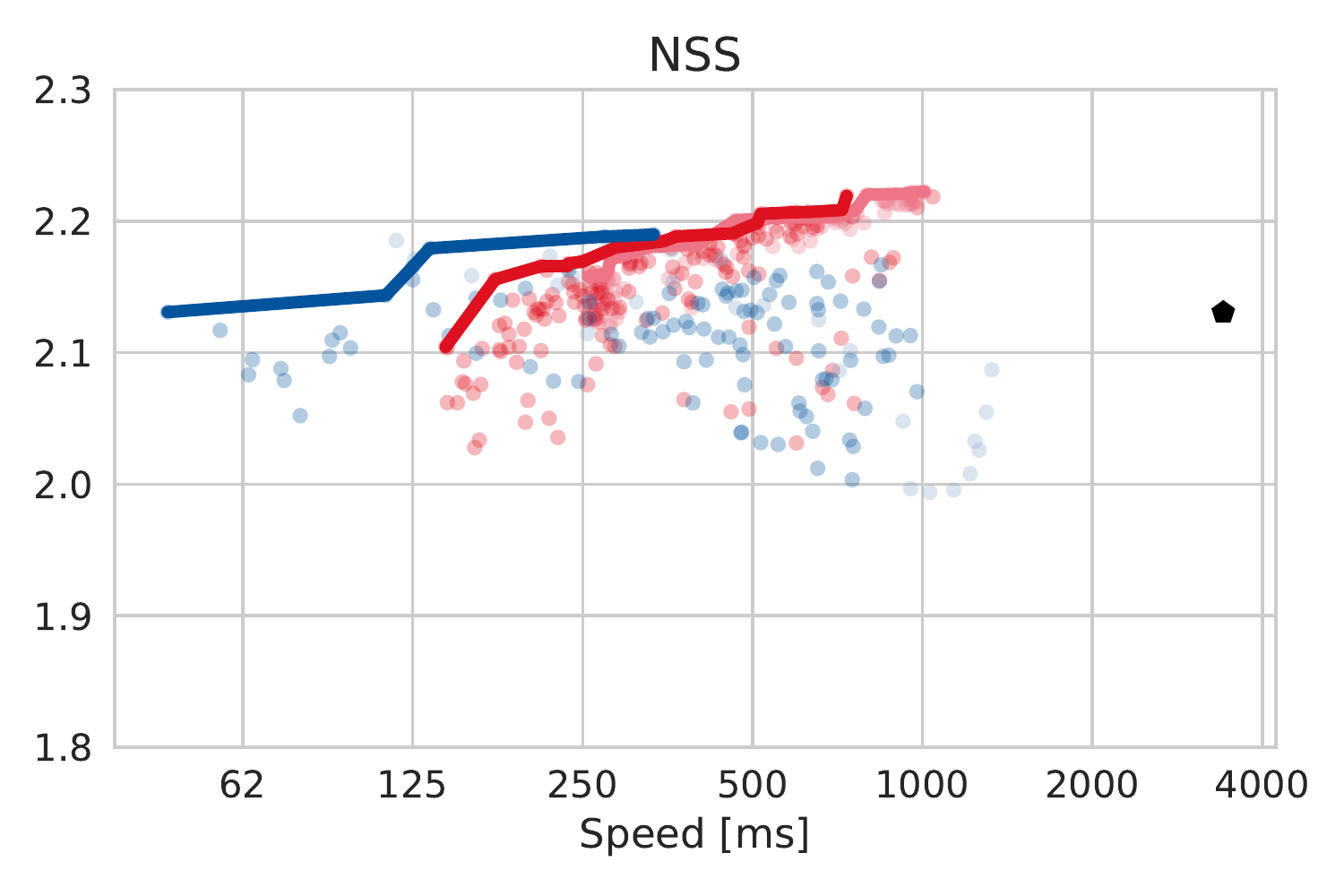}
      \caption{Speed and performance of various models trained on MIT1003 and evaluated
      on the CAT2000 training set. Each point corresponds to a different architecture
      with a different number of pruned features and a different $\beta$.
      Solid lines mark Pareto optimal models. In some cases, only a single model is Pareto optimal
      and no solid line is shown. The $x$-axis shows speed measured on a CPU for a single image
      as input. FastGaze reaches faster speeds than DenseGaze, especially when explicitly
      optimized to reduce the amount of floating point operations ($\beta > 0$). With the exception
      of SIM, DenseGaze reaches higher prediction performance. For all metrics, even heavily pruned
      models appear to generalize as well or better to CAT2000 than DeepGaze~II.}
      \label{fig:cpu_vs_auc}
    \end{figure*}

    In the following, we first validate the performance of Fisher pruning on a simpler toy example.
    We then explore the performance and computational cost of two architectures for saliency prediction.
    First, we try using the smaller VGG-11 variant of Simonyan et al. \cite{simonyan2014vgg} for feature extraction.
    In contrast to the readout network of
    K{\"u}mmerer et al. \cite{kummerer2016deepgaze2}, which
    took as input feature maps from 5 different layers, we only used the output of the last convolutional layer
    (``conv5\_2'') as input. Extracting information from multiple layers is less important in our
    case, since we are also optimizing the parameters of the feature extraction network. As an
    alternative to VGG, we try DenseNet-121 as a feature extractor
    \cite{huang2017densely}, using the output of ``dense block 3'' as input to the readout network.
    In both cases, the readout network consisted of convolutions with parametric rectified linear
    units \cite{he2015prelu} and 32, 16, and 2 feature maps in the hidden layers.
    In the following, we will call the first network \textit{FastGaze} and the second network
    \textit{DenseGaze}.

    \subsection{Fisher Pruning}
    \begin{table}[t]
      \centering
      \caption{Comparison of pruning methods for the LeNet-5 network trained on MNIST~\cite{lecun1998lenet}}
      \begin{tabular}{p{2.5cm}|cc}
        \hline
        Method & Error & Computational cost \\
        \hline
        LeCun et al. \cite{lecun1998lenet} & 0.80\% & 100\% \\
        Han et al. \cite{han2015pruning} & 0.77\% &16\% \\
        \textbf{Fisher ($\beta = 0$)} & 0.84\% & 26\% \\
        \textbf{Fisher ($\beta > 0$)} & 0.79\% & \textbf{10\%} \\
        \textbf{Fisher ($\beta^*$)} & 0.86\% & 17\% \\
        \hline
      \end{tabular}
      \label{tbl:lenet}
    \end{table}

    We apply Fisher pruning to the example of a LeNet-5 network trained on the MNIST dataset \cite{lecun1998lenet}.
    We compare our method to the pruning method of Han et al. \cite{han2015pruning} which requires $L_1$ or $L_2$ regularization of the model's parameters during
    an initial training phase and cannot directly be applied to a pretrained model. LeNet-5 consists
    of two convolutional layers and two pooling layers, followed by two fully connected layers. Following Han et
    al. \cite{han2015pruning}, the details of the initial architecture were the same as in the MNIST
    example provided by the Caffe framework \footnote{https://github.com/BVLC/caffe/}. We used 53000
    data points of the training set for training and 7000 data points for validation and early
    stopping.

    We find that Fisher pruning performs well, but that taking the computational cost into account is important
    (Table~\ref{tbl:lenet}). Automatically choosing the weight controlling the trade-off between performance and
    computational cost works better than ignoring computational cost $(\beta^*)$, although not as well as
    using a fixed but optimized weight $(\beta > 0)$.

    An incorrect decision of a pruning algorithm may be corrected by retraining the
    network's weights, especially for toy examples like LeNet-5 which are quick to train.
    This can mask the bad performance of a poor pruning algorithm. In Figure~\ref{fig:pruning_comparison},
    we therefore look at the pruning performance of various pruning techniques applied to FastGaze
    while keeping the parameters fixed. We alternate between estimating the pruning signal on the
    full MIT1003 training set and pruning 10 features at a time.
    We included the method of Molchanov et al. \cite{molchanov2017pruning} as well as two naive
    baselines, namely pruning based on the average activity of a feature, $\frac{1}{N \cdot H \cdot W} \sum_{nij}
    |a_{nkij}|$ (L1A), and pruning based on the average $L_1$-norm of a weight vector corresponding
    to a feature map, $||\mathbf{w}_k||_1$ (L1W).

    While the simple baselines are intuitive (i.e., if a feature is ``off'' most of the time, one might expect it to
    not contribute much to a network's performance), they are also inherently flawed. The
    activations and parameters of a layer with ReLU activations can be arbitrarily scaled without changing
    the function implemented by the network by compensating for the scaling in the parameters of the
    next layer. Correspondingly, we find that the baselines perform poorly.

    We further find that unregularized Fisher pruning performs as well as the method of
    Molchanov et al. \cite{molchanov2017pruning}, independent of whether we use the normalized or
    unnormalized variant of their pruning signal. However, we find that our regularization of the
    number FLOPs gives better results. Using different $\beta$ for differently strongly pruned
    networks appears to be important, as well as updating the cost of a feature map as neighboring
    feature maps get pruned (Figure~\ref{fig:pruning_comparison}).

    Finally, we also tested our alternative pruning signal which automatically tunes the trade-off
    weight ($\beta^*$). The results suggest that this approach works well when the number of features
    to be pruned is small, but may work worse when the number of pruned features is large (Figure~\ref{fig:pruning_comparison}).

    \subsection{Pruning FastGaze and DenseGaze}

    \begin{figure*}[bt]
      \centering
      \begin{tabular}{m{0.2\textwidth}m{0.5\textwidth}m{0.2\textwidth}}
        \includegraphics[trim={9cm 0cm 9cm 0cm},clip,width=0.2\textwidth]{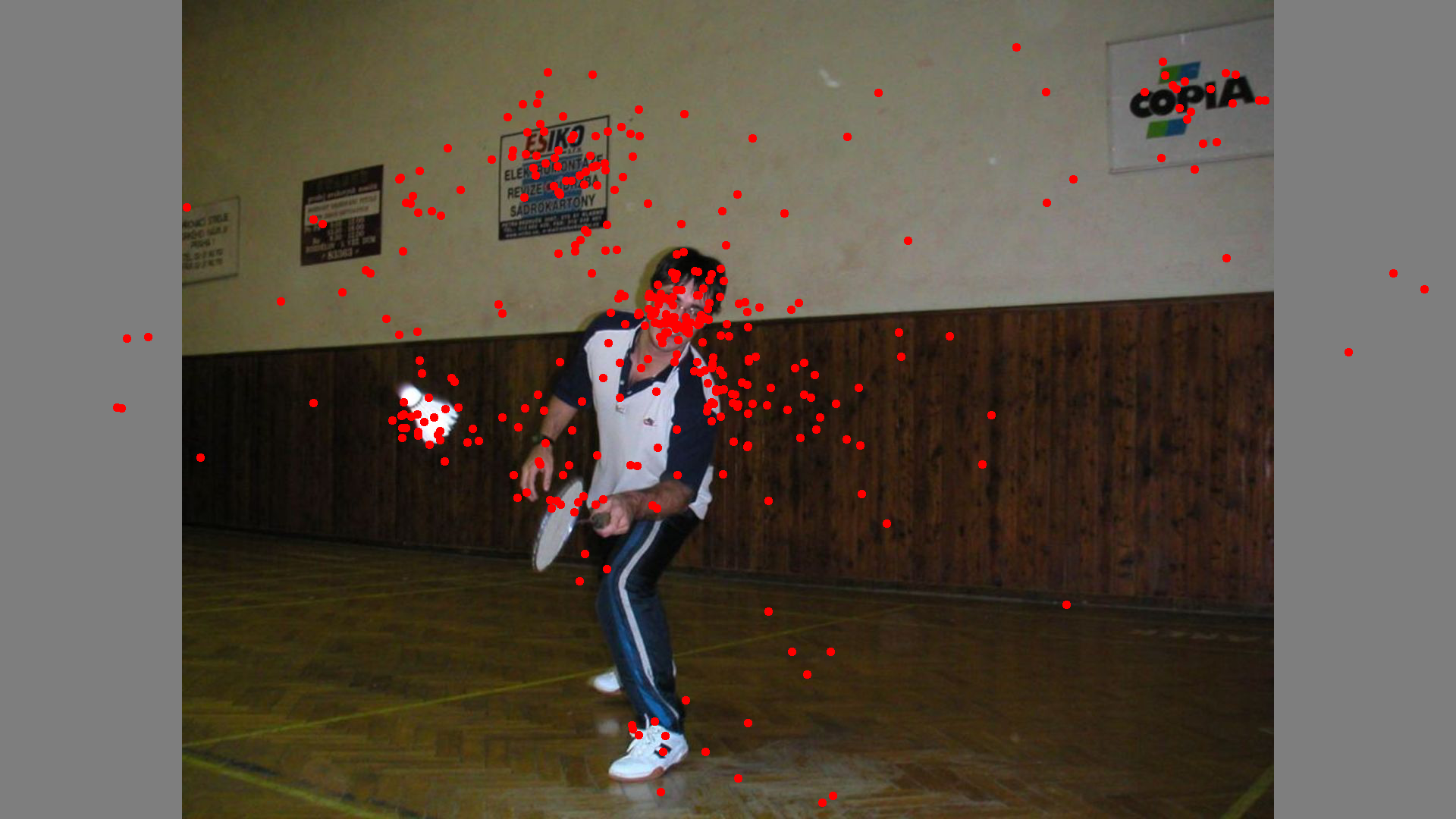}
        & \includegraphics[trim={5cm 2cm 4.5cm 2cm},clip,width=0.5\textwidth]{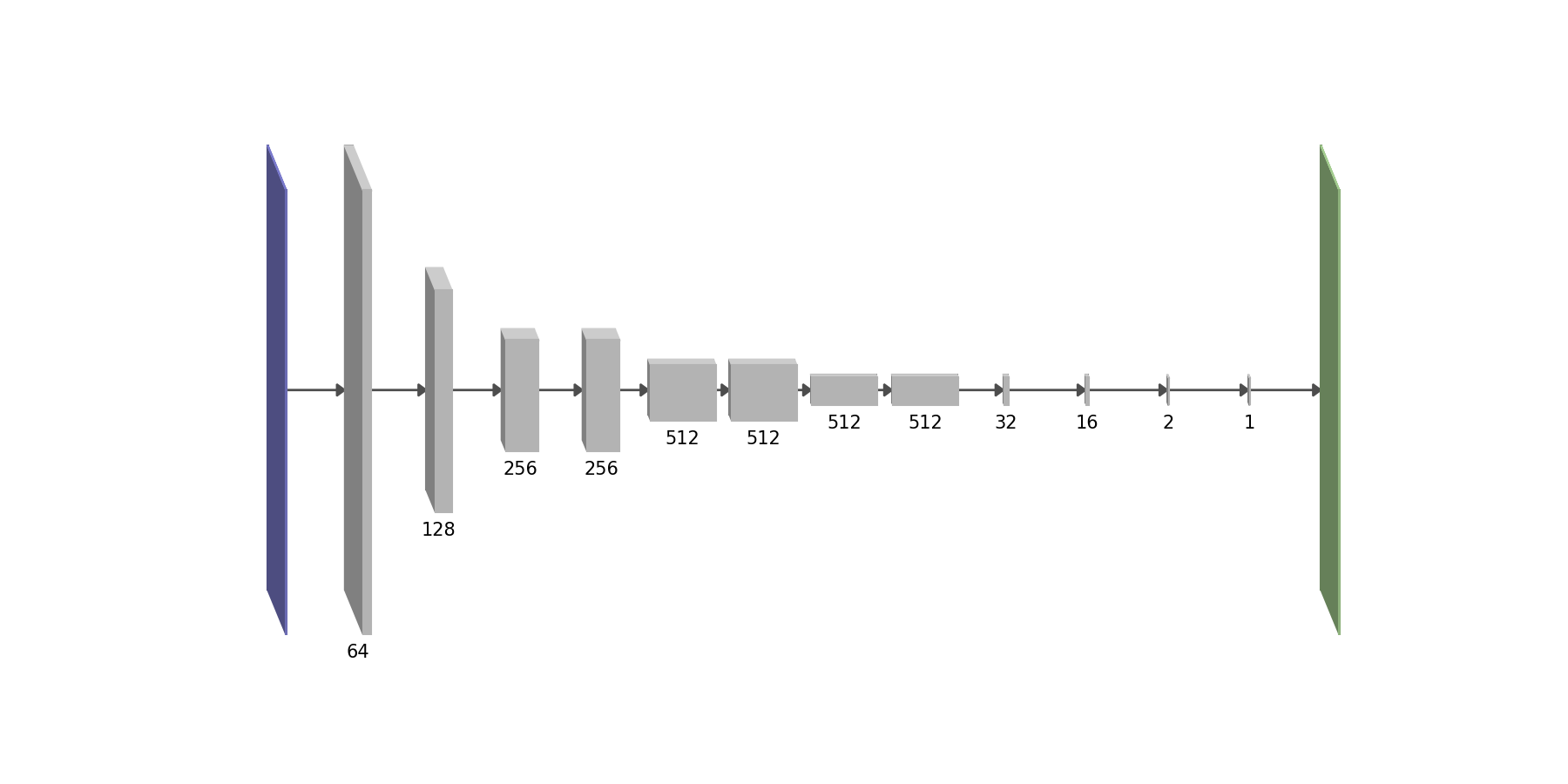} &
        \includegraphics[trim={9cm 0cm 9cm 0cm},clip,width=0.2\textwidth]{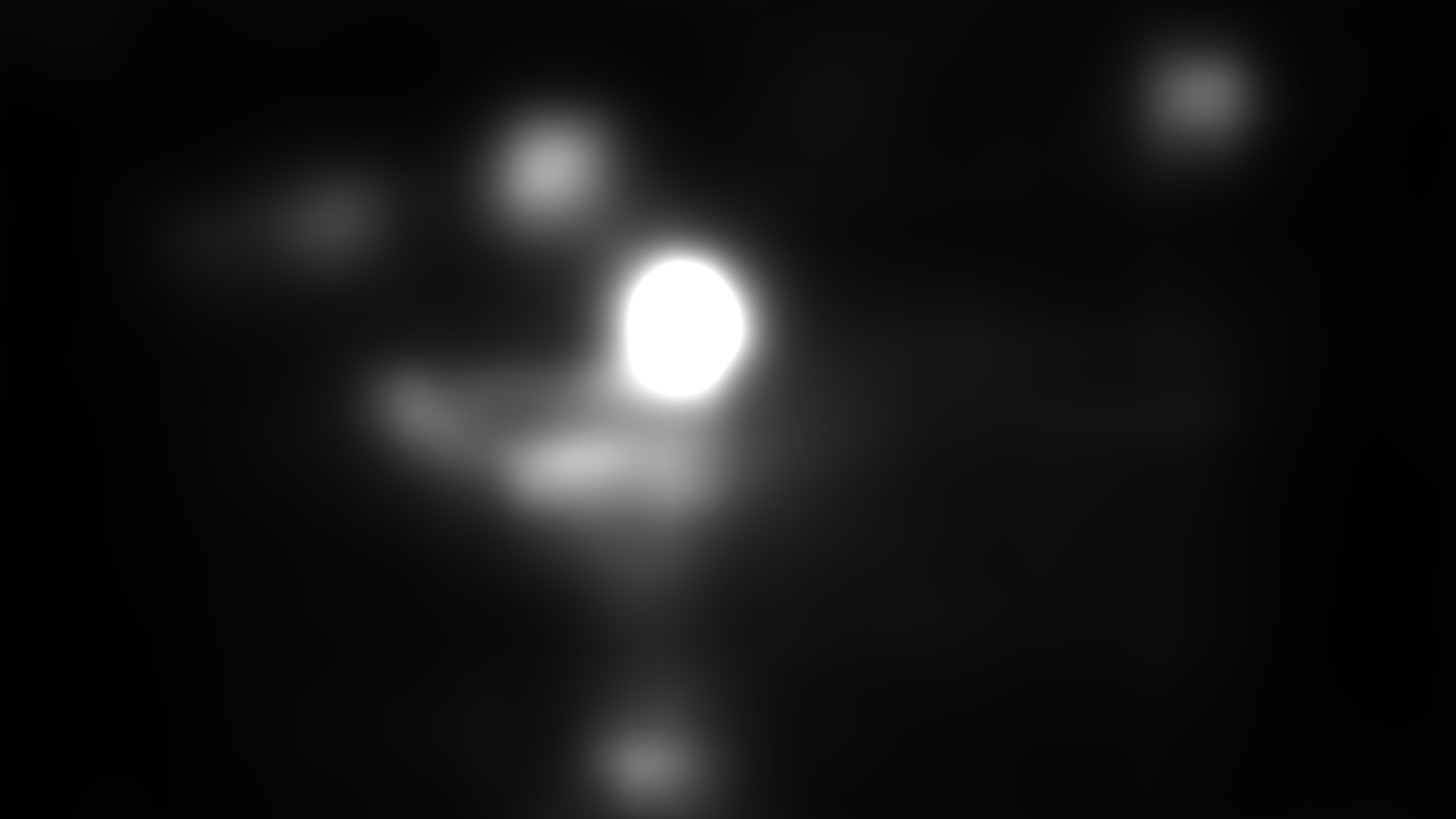} \\
        & \includegraphics[trim={5cm 2cm 4.5cm 2cm},clip,width=0.5\textwidth]{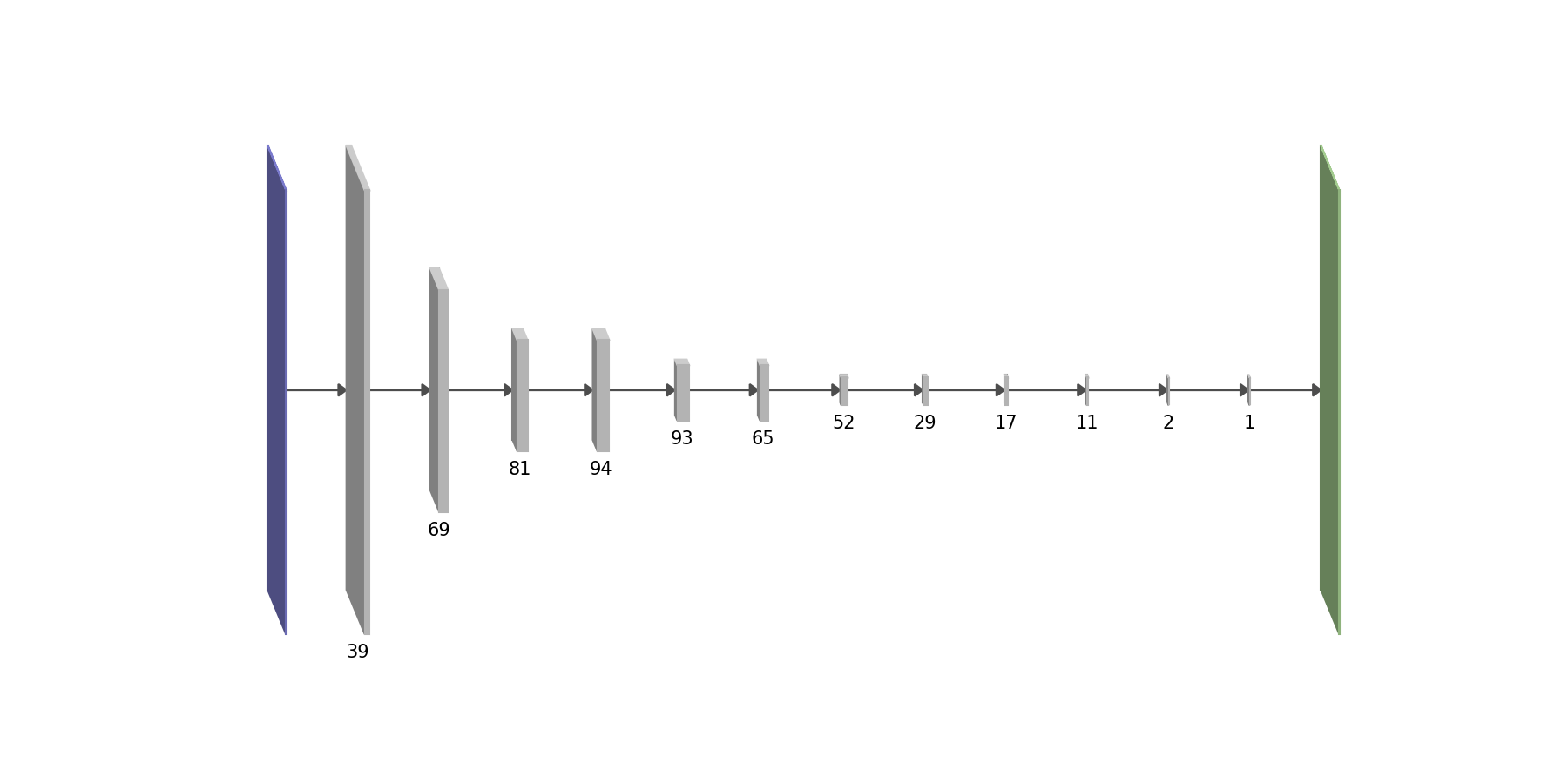} &
        \includegraphics[trim={9cm 0cm 9cm 0cm},clip,width=0.2\textwidth]{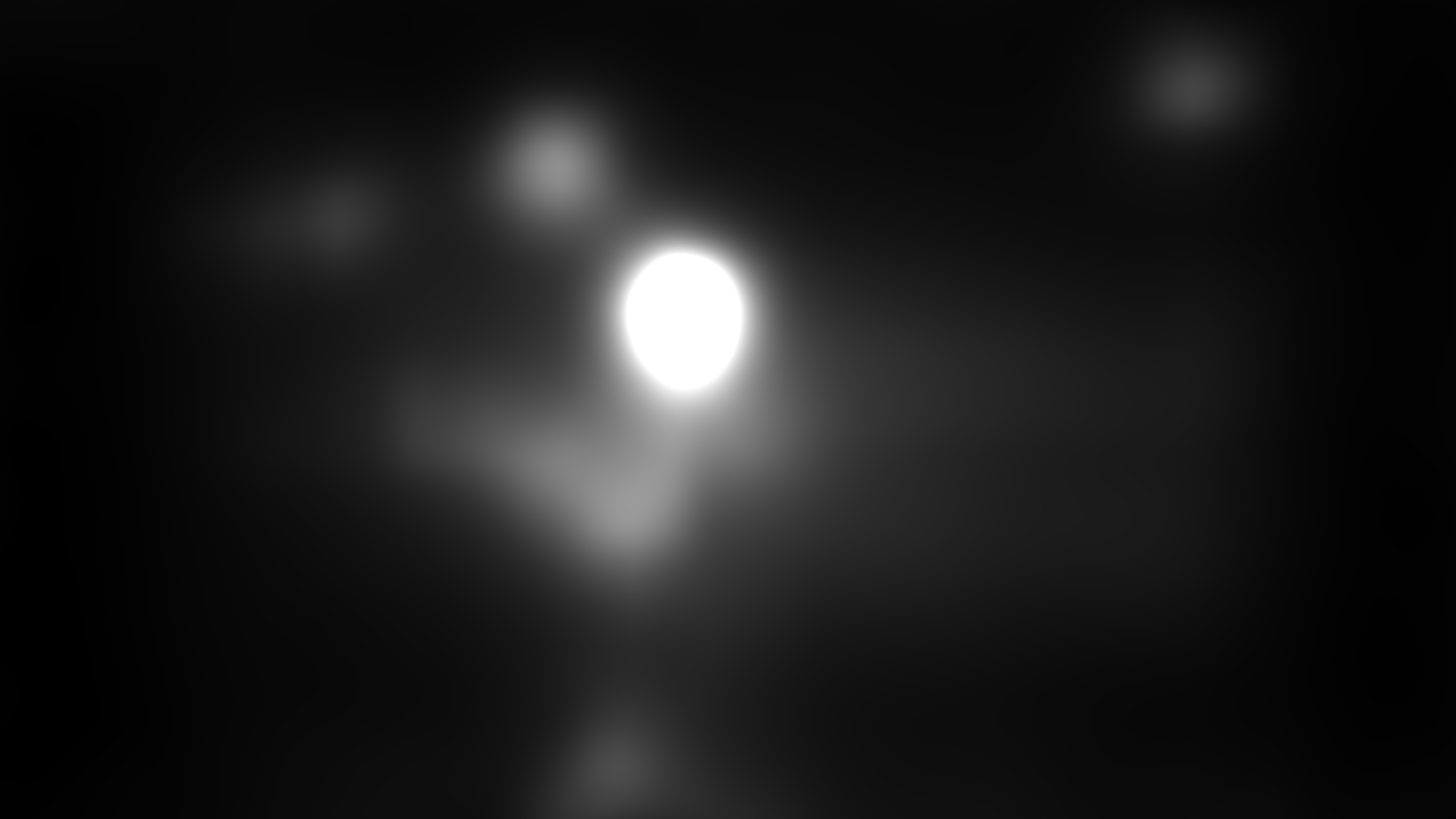} \\
        & \includegraphics[trim={5cm 2cm 4.5cm 2cm},clip,width=0.5\textwidth]{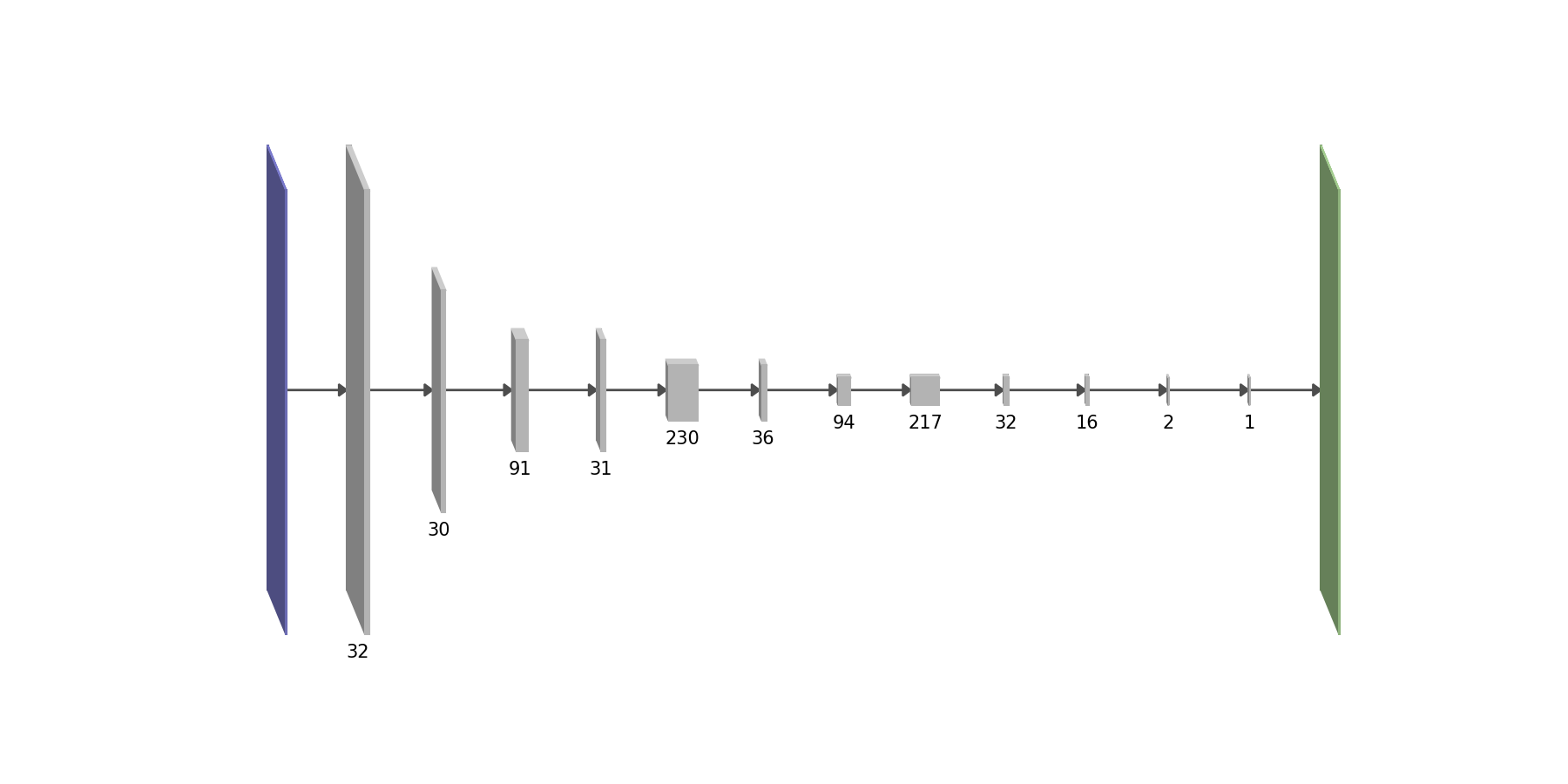} & 
        \includegraphics[trim={9cm 0cm 9cm 0cm},clip,width=0.2\textwidth]{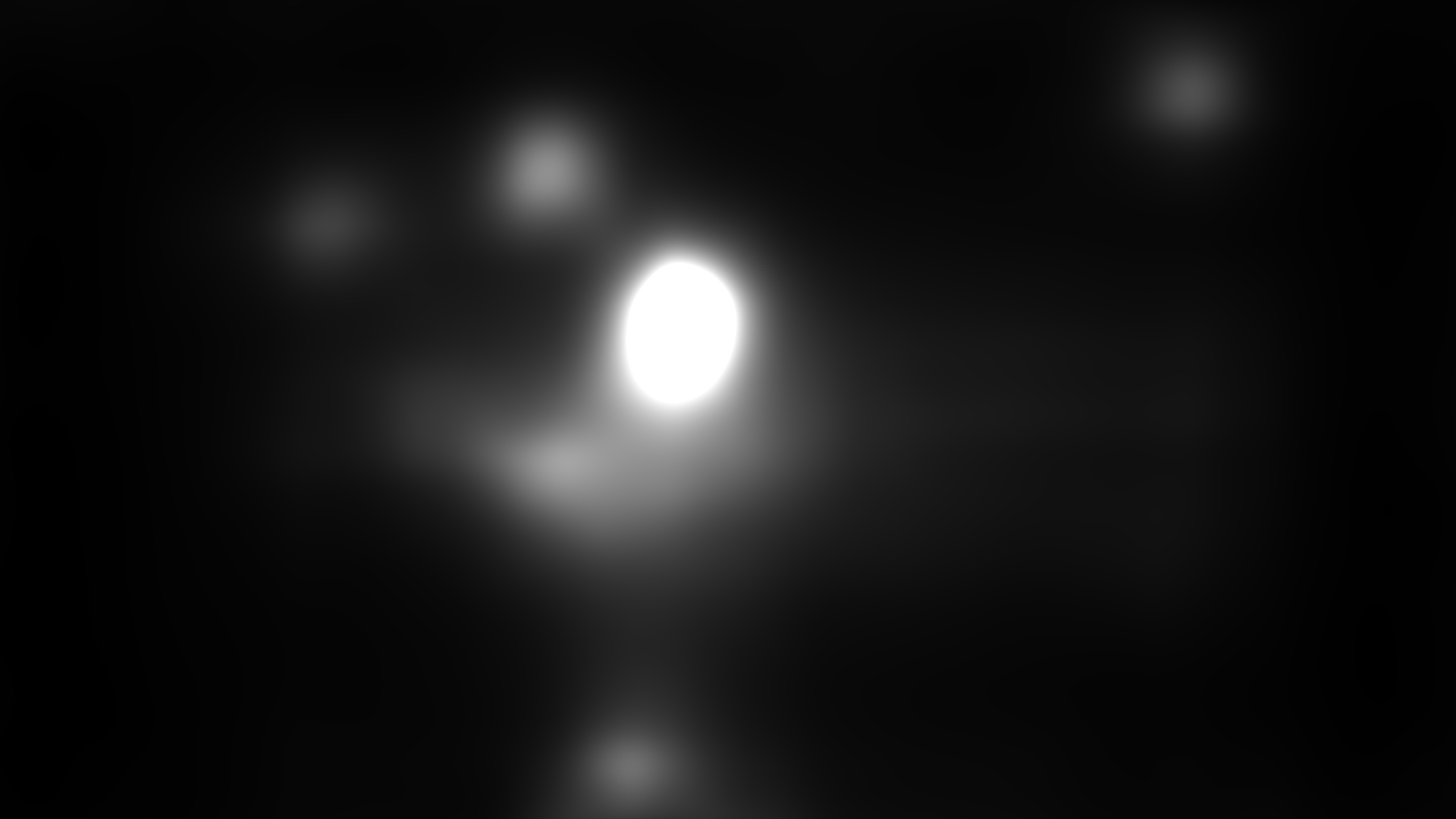} \\
        & \includegraphics[trim={5cm 2cm 4.5cm 2cm},clip,width=0.5\textwidth]{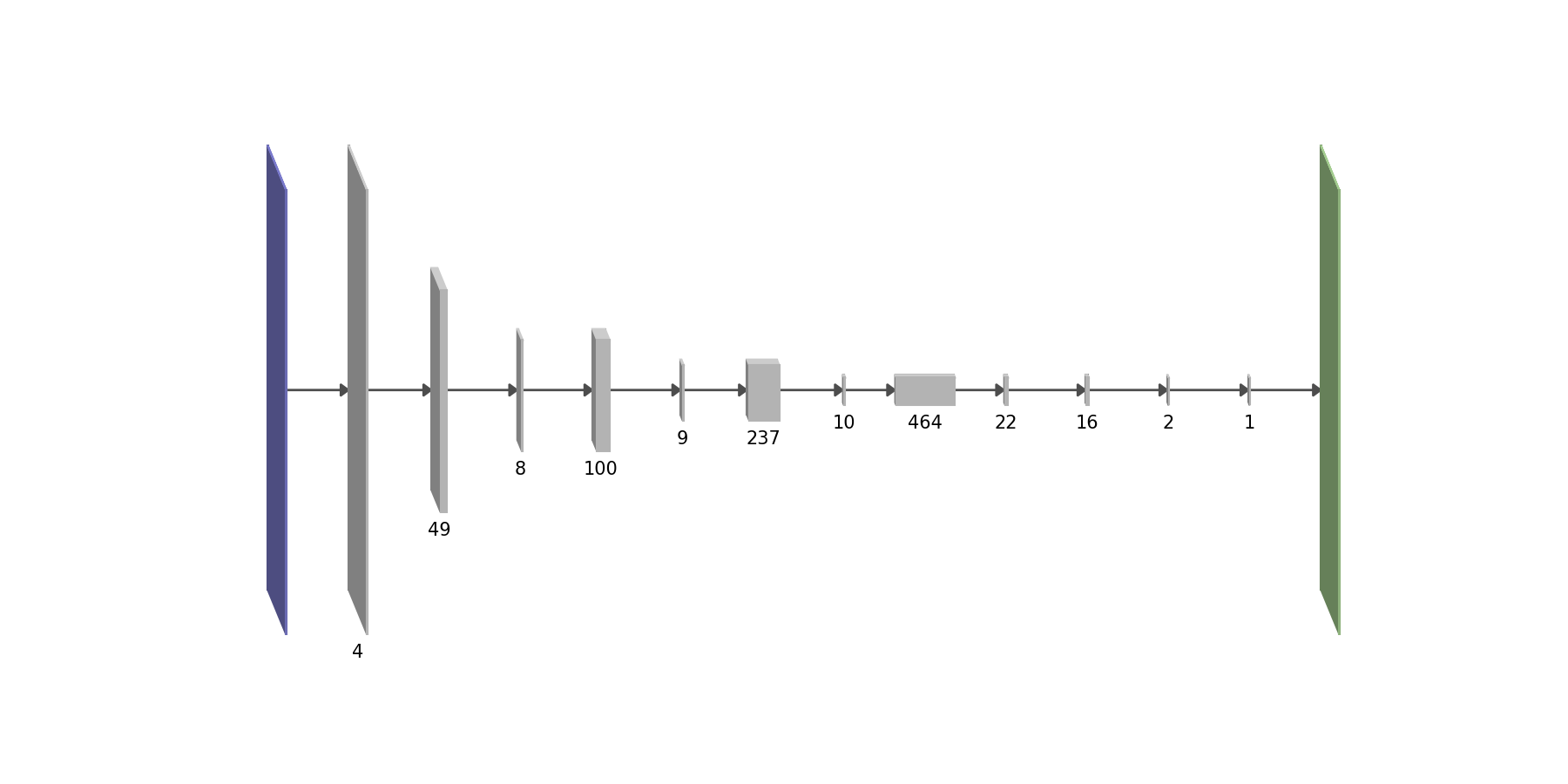} &
        \includegraphics[trim={9cm 0cm 9cm 0cm},clip,width=0.2\textwidth]{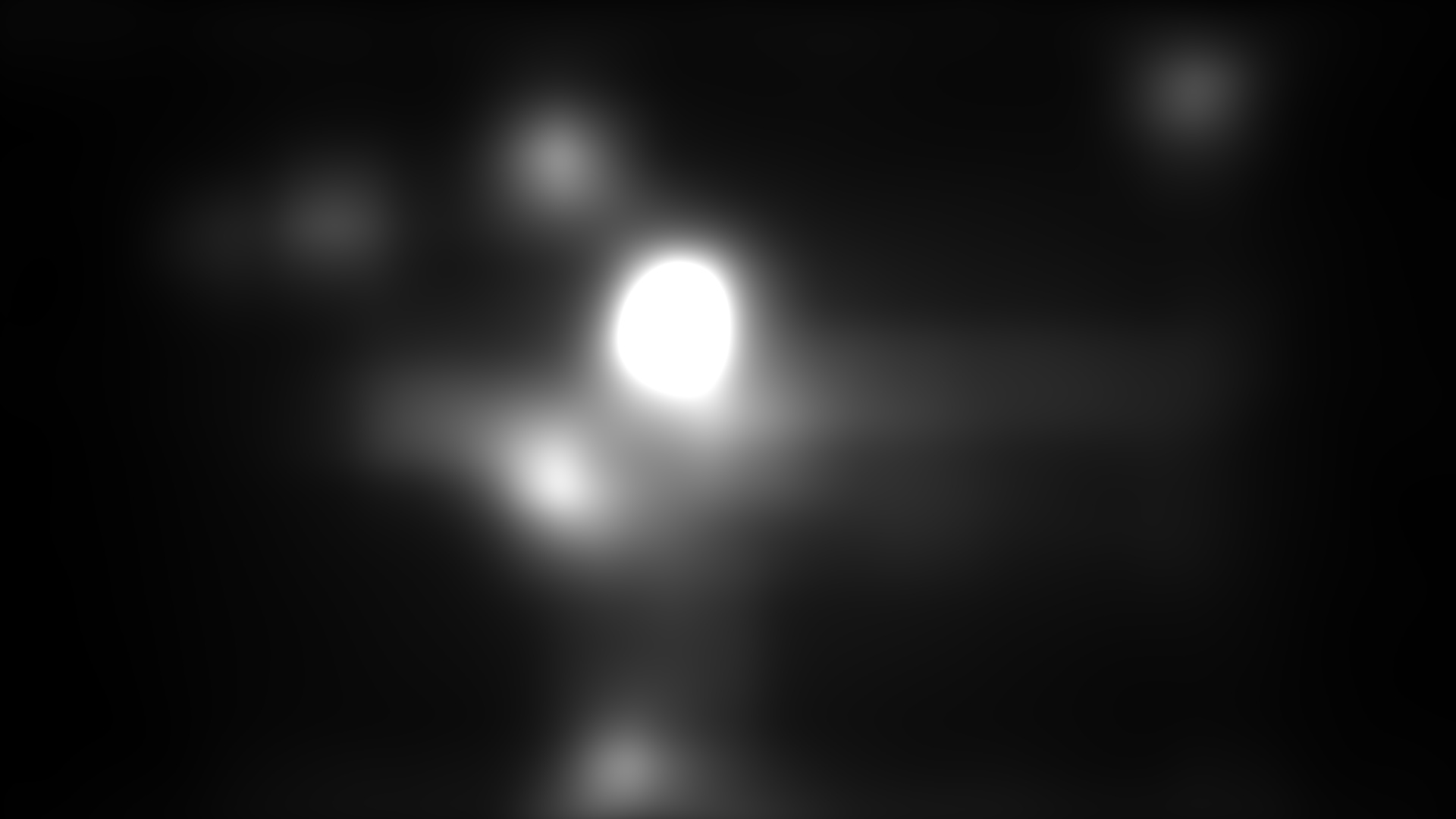}
      \end{tabular}
      \caption{The top graph visualizes the feature maps of the unpruned FastGaze model, while the remaining graphs
      visualize models pruned to different degrees. Labels indicate the number of feature maps. From top to bottom, the measured runtime for these
      models was 1.39s, 356ms, 258ms, and 91ms. For strongly regularized models, we observed a tendency to alternate
      between convolutions with many feature maps and convolutions with very few feature maps. The top left corner shows an example input image
      and ground truth fixations, while the right column shows saliency models generated by the different models. Despite a 15x difference in speed,
      the saliency maps are visually very similar.}
      \label{fig:pruned_models}
    \end{figure*}

    To find the optimal pruning parameters, we ran multiple experiments with a randomly sampled
    number of pruned features and a randomly chosen trade-off parameter $\beta$. The total number of
    feature maps is 2803 for FastGaze and 7219 for DenseGaze. $\beta$ was chosen between 3e-4 and
    3e-1. We evaluated each model in terms of log-likelihood, area under the curve (AUC), \textit{normalized scanpath
    saliency} (NSS), and \textit{similarity} (SIM). Other metrics commonly used in saliency
    literature such as sAUC or CC are closely related to one of these metrics \cite{Kuemmerer2017a}.
    We used the publicly available CAT2000 \cite{borji2015cat2000} dataset for evaluation, which was not used during
    training of any of the models. While there are many ways to measure computational cost, we here were mostly interested in single-image
    performance on CPUs. We used a single core of an Intel Xeon E5-2620 (2.4GHz) and averaged the
    speed over 6 images of 384 x 512 pixels. Our implementation was written in PyTorch \cite{PyTorch}.

    Figure~\ref{fig:cpu_vs_auc} shows the performance of various models. In terms of log-likelihood,
    NSS, and SIM, we find that both FastGaze and DenseGaze generalize better to CAT2000 than our reimplementation
    of DeepGaze~II, despite the fact that both models were regularized to imitate DeepGaze~II. In terms
    of AUC, DeepGaze~II performs slightly better than FastGaze but is outperformed by DenseGaze.
    Pruning only seems to have a small effect on performance, as even heavily pruned models still perform well.
    For the same AUC, we achieve a speedup of
    roughly 10x with DenseGaze, while in terms of log-likelihood even our most heavily pruned model yielded
    better performance (which corresponds to a speedup of more than 75x).
    Comparing DenseGaze and FastGaze, we find that while DenseGaze achieves better AUC performance, 
    FastGaze is able to achieve faster runtimes due to its less complex architecture.

    \begin{figure*}[bt]
      \begin{tabular}{cccccc}
        \rotatebox{90}{\hspace{.45cm}\textsf{\textbf{Data}}} &
        \includegraphics[trim={8cm 0cm 8cm 0cm},clip,height=0.14\textwidth]{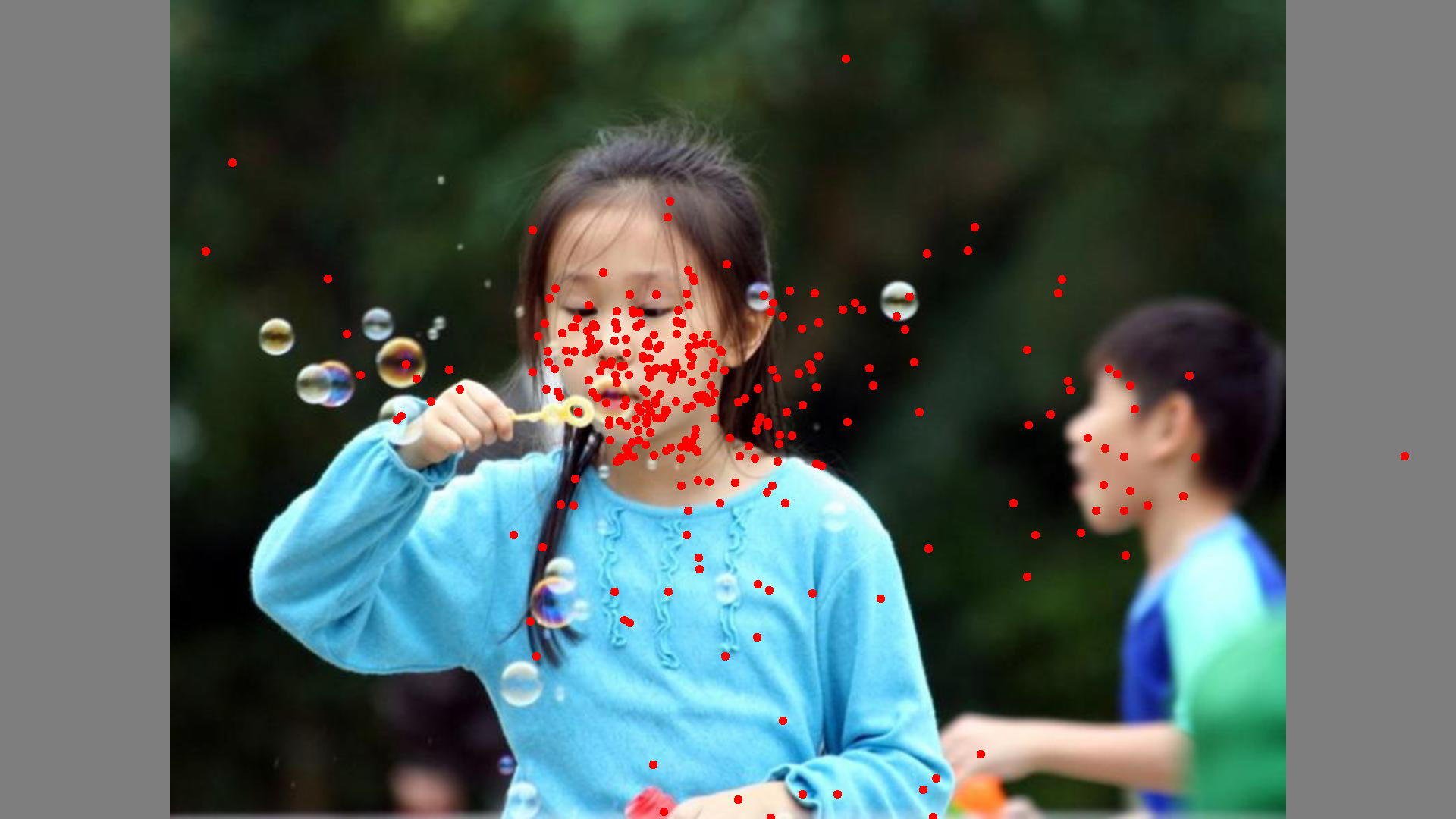} &
        \includegraphics[trim={3cm 0cm 3cm 0cm},clip,height=0.14\textwidth]{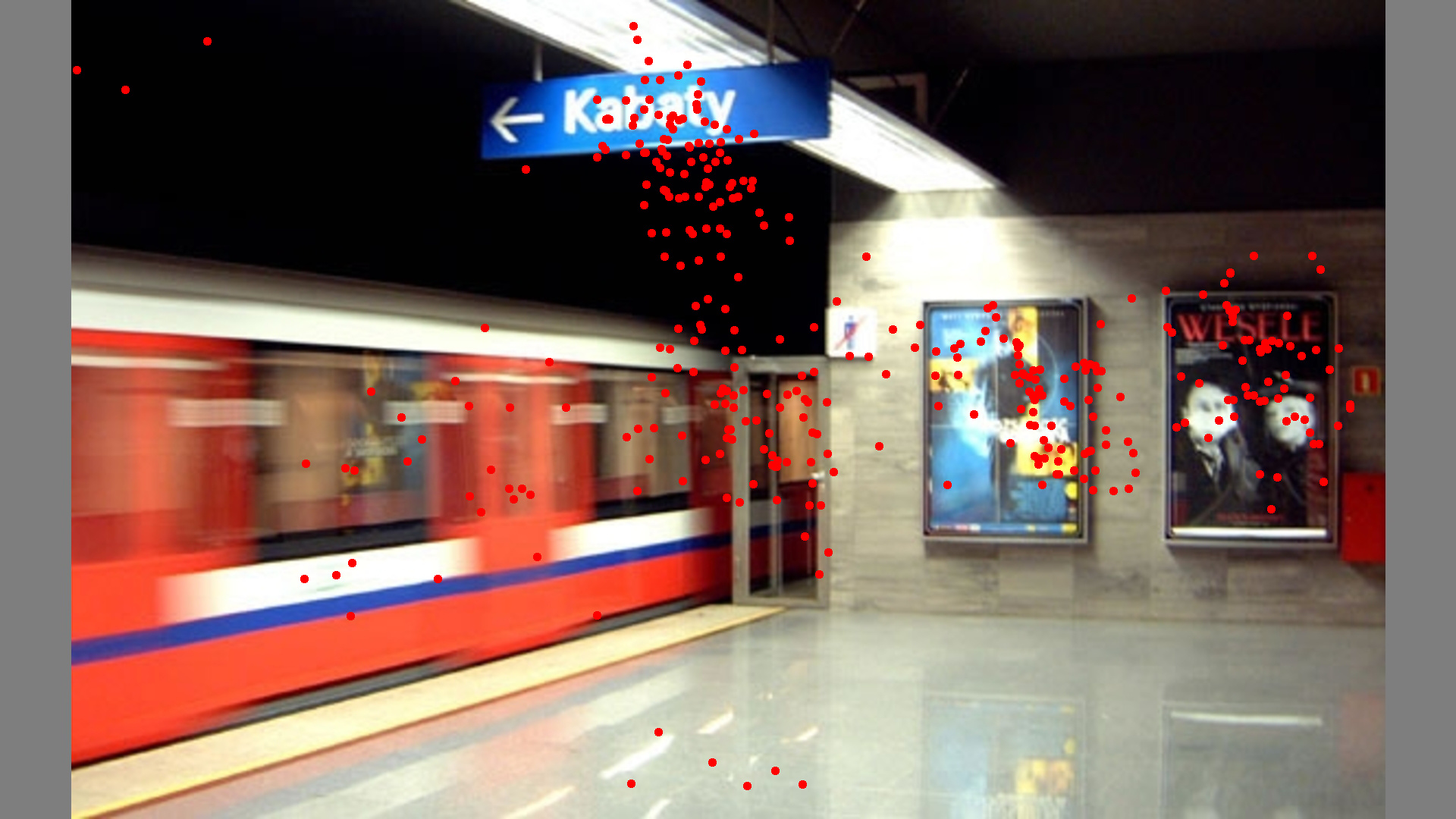} &
        \includegraphics[trim={9cm 0cm 9cm 0cm},clip,height=0.14\textwidth]{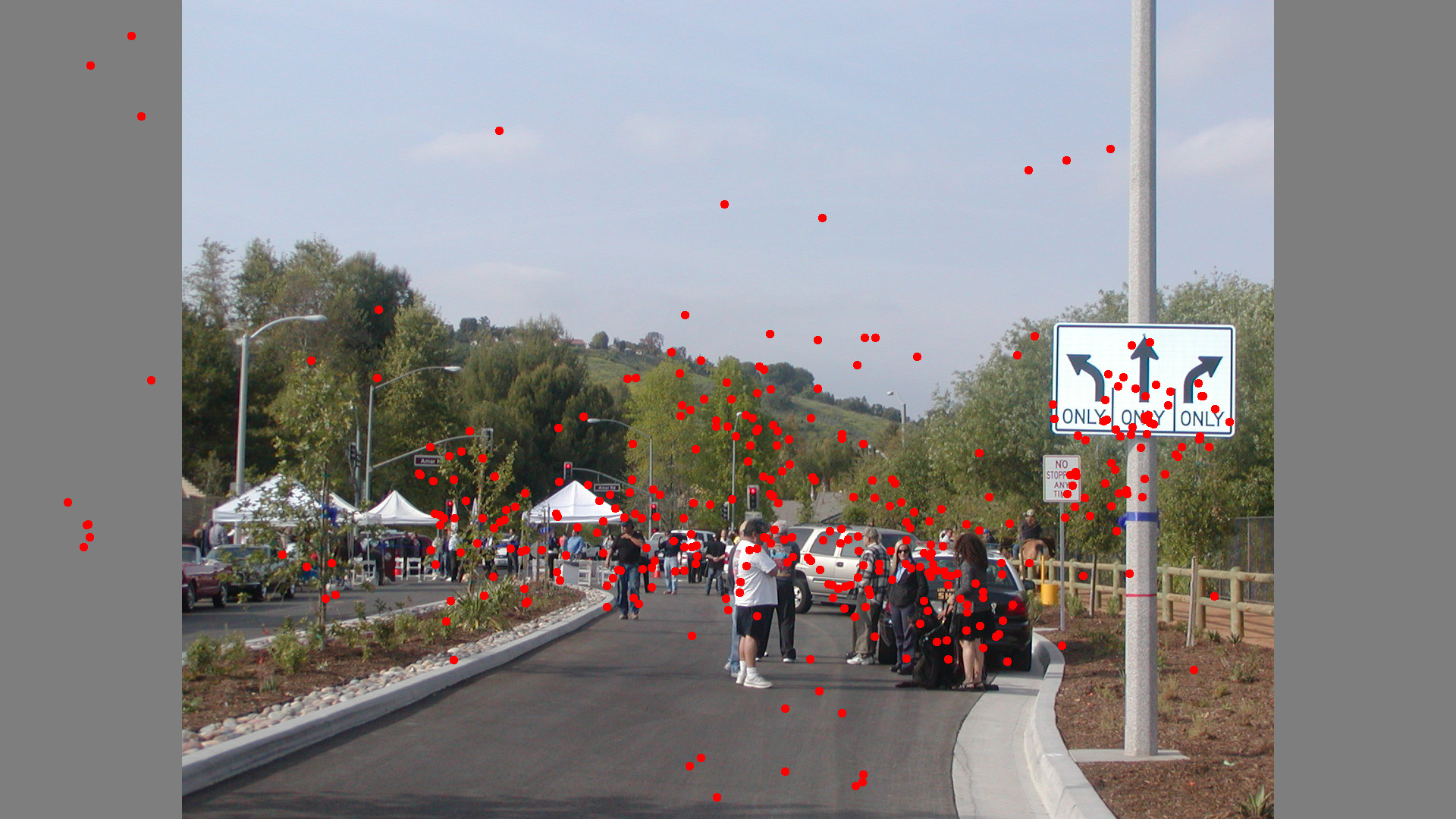} &
        \includegraphics[trim={18cm 0cm 18cm 0cm},clip,height=0.14\textwidth]{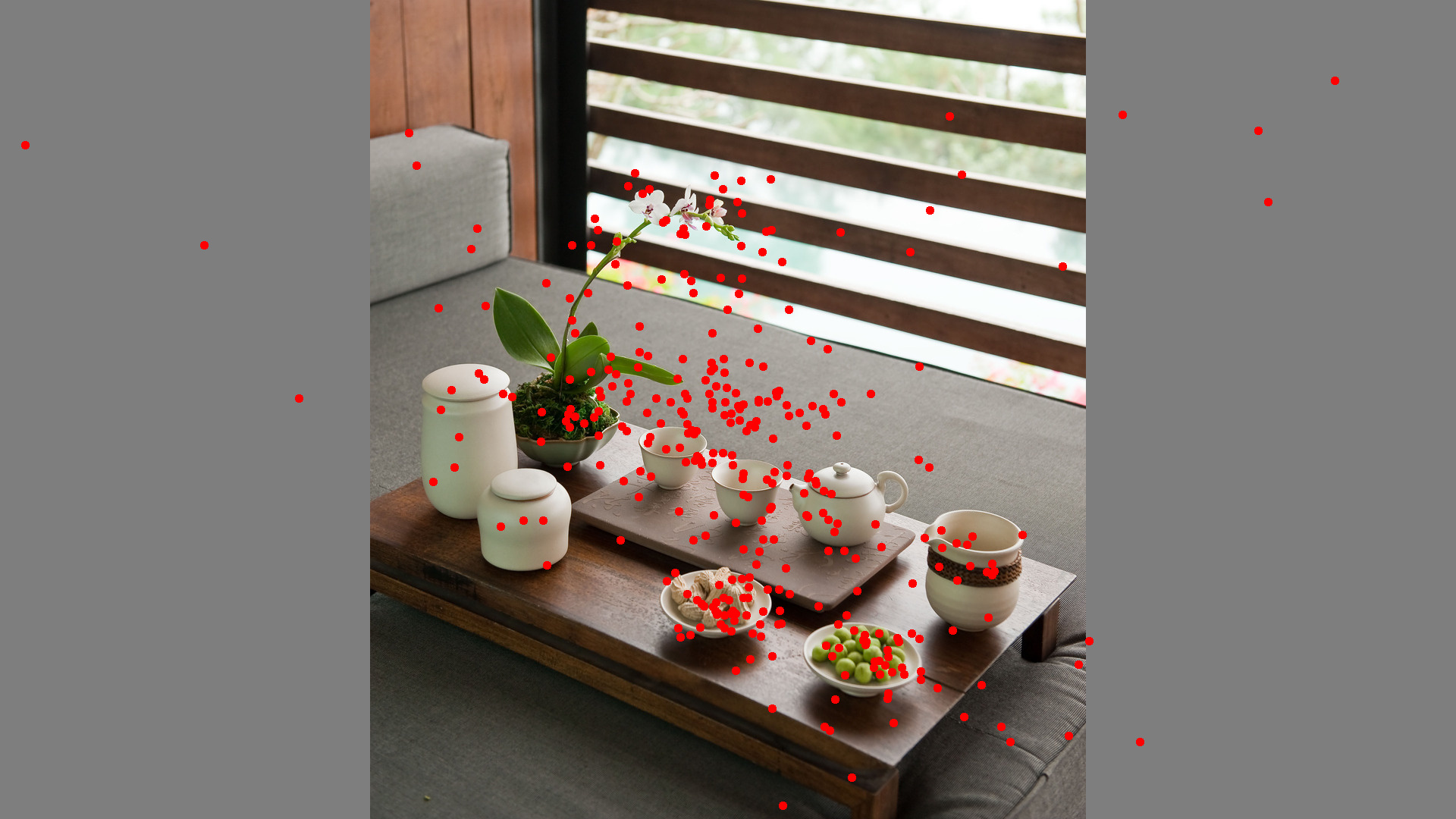} &
        \includegraphics[trim={9cm 0cm 9cm 0cm},clip,height=0.14\textwidth]{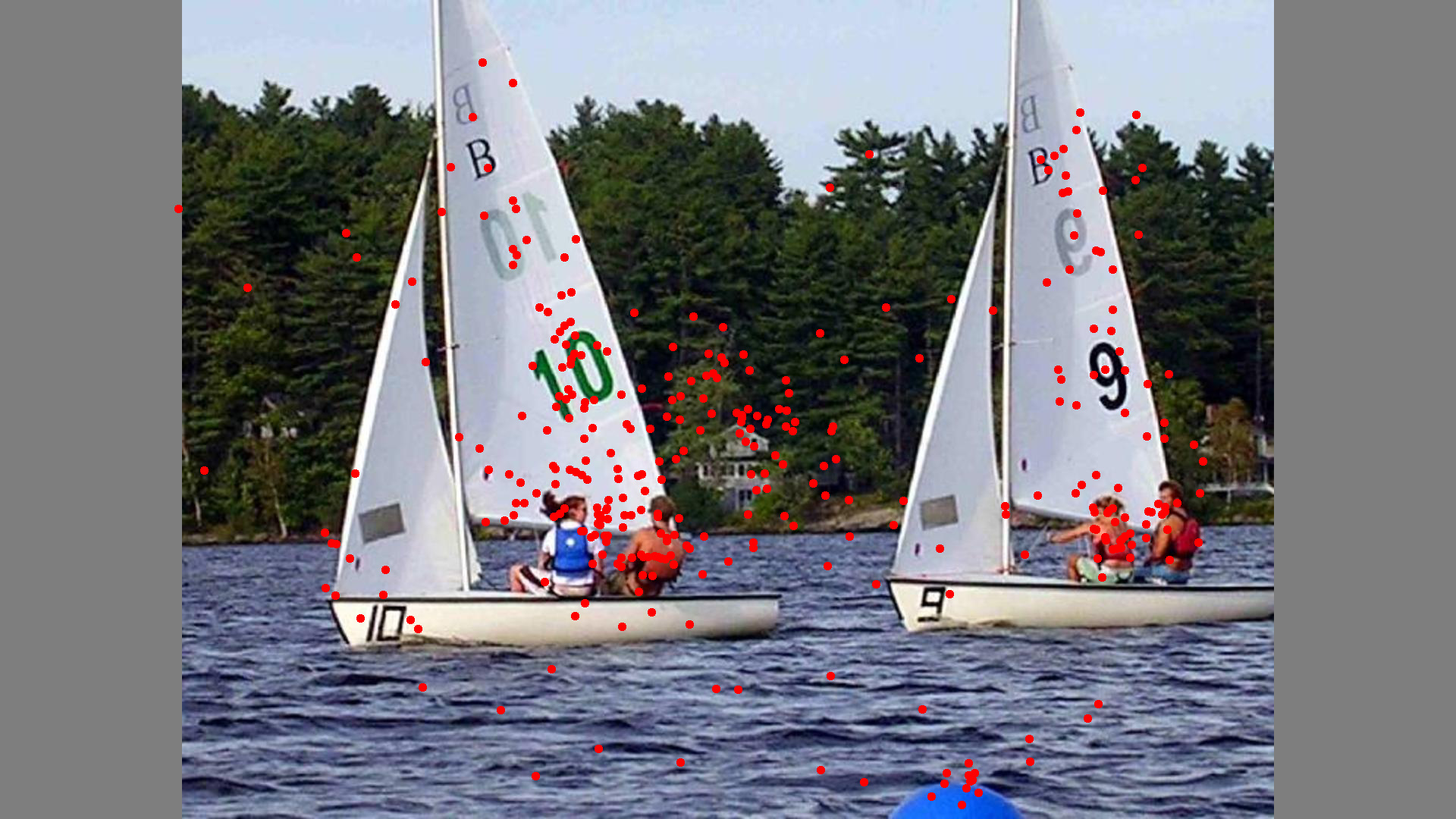} \\
        \rotatebox{90}{\hspace{-.03cm}\textsf{\textbf{DeepGaze~II}}} \rotatebox{90}{\hspace{.4cm}(3.59s)} &
        \includegraphics[trim={8cm 0cm 8cm 0cm},clip,height=0.14\textwidth]{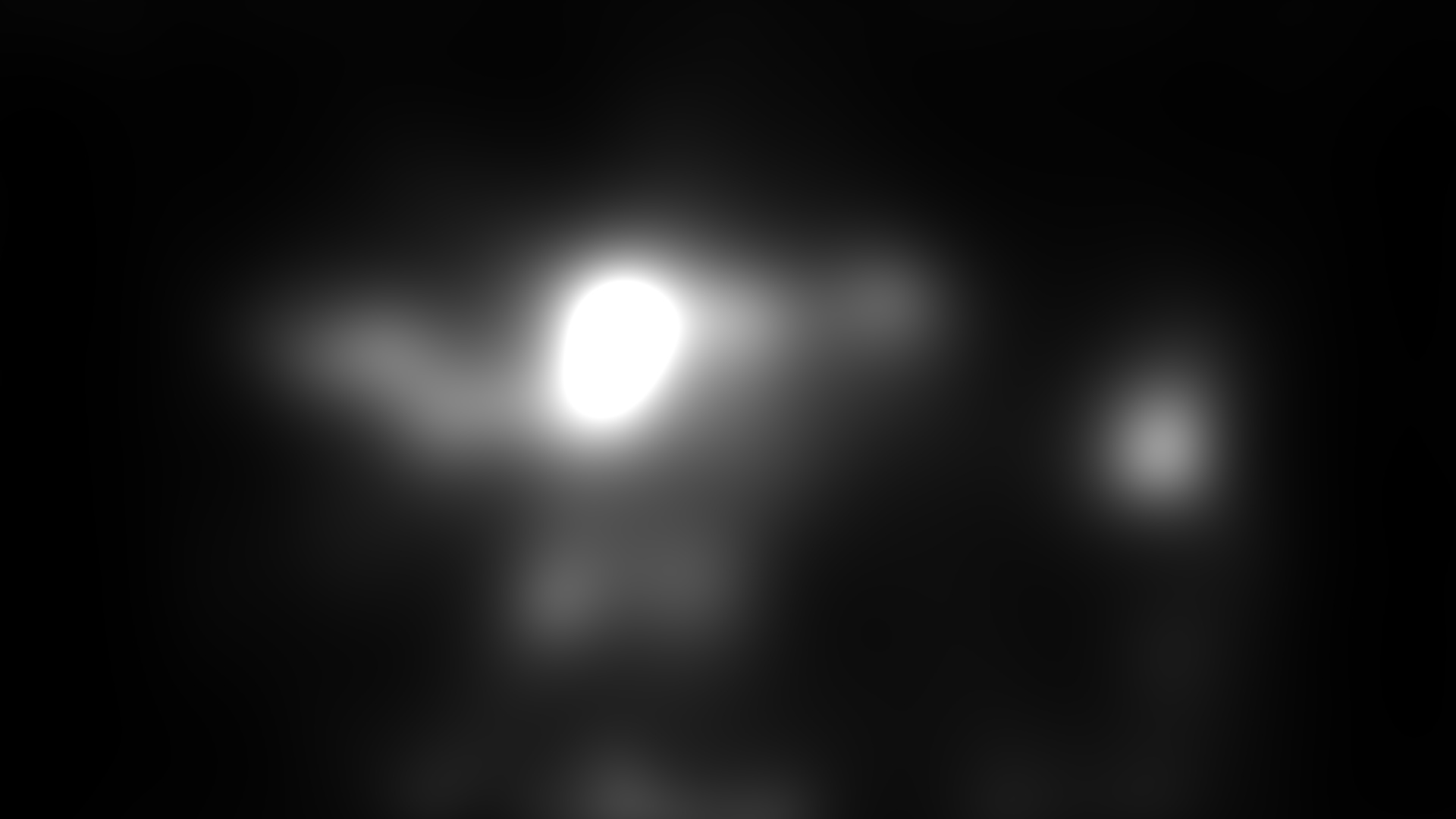} &
        \includegraphics[trim={3cm 0cm 3cm 0cm},clip,height=0.14\textwidth]{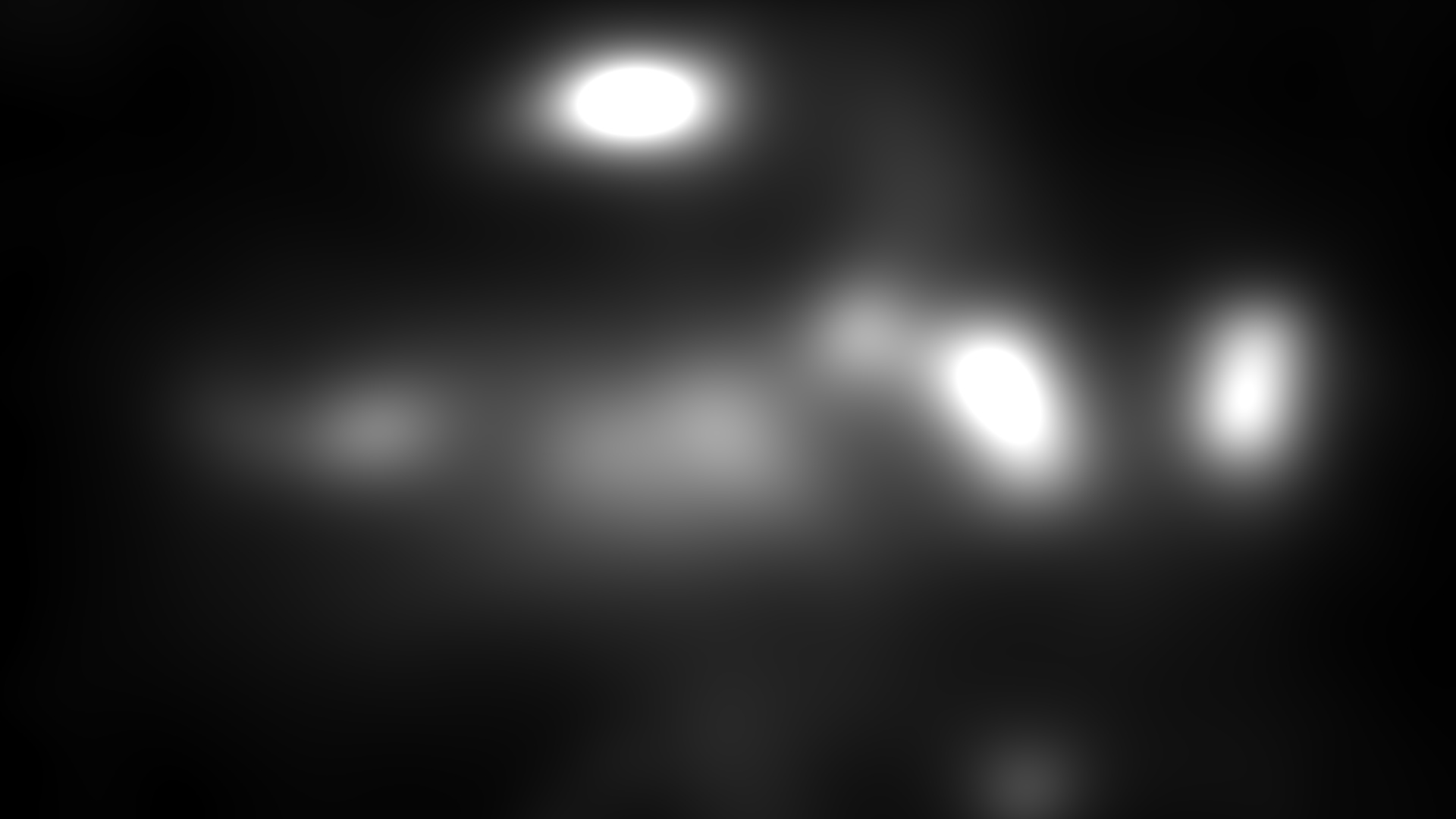} &
        \includegraphics[trim={9cm 0cm 9cm 0cm},clip,height=0.14\textwidth]{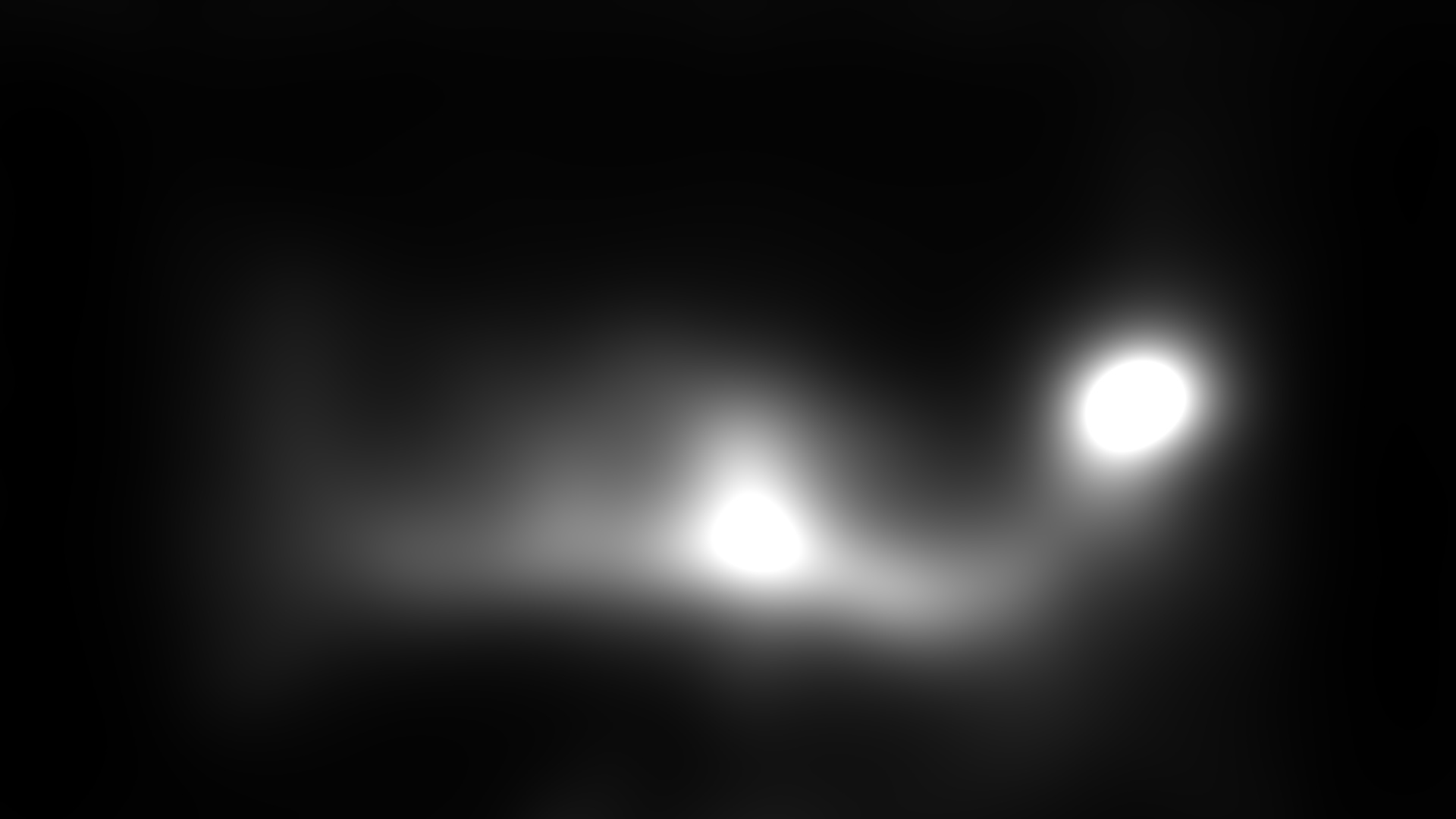} &
        \includegraphics[trim={18cm 0cm 18cm 0cm},clip,height=0.14\textwidth]{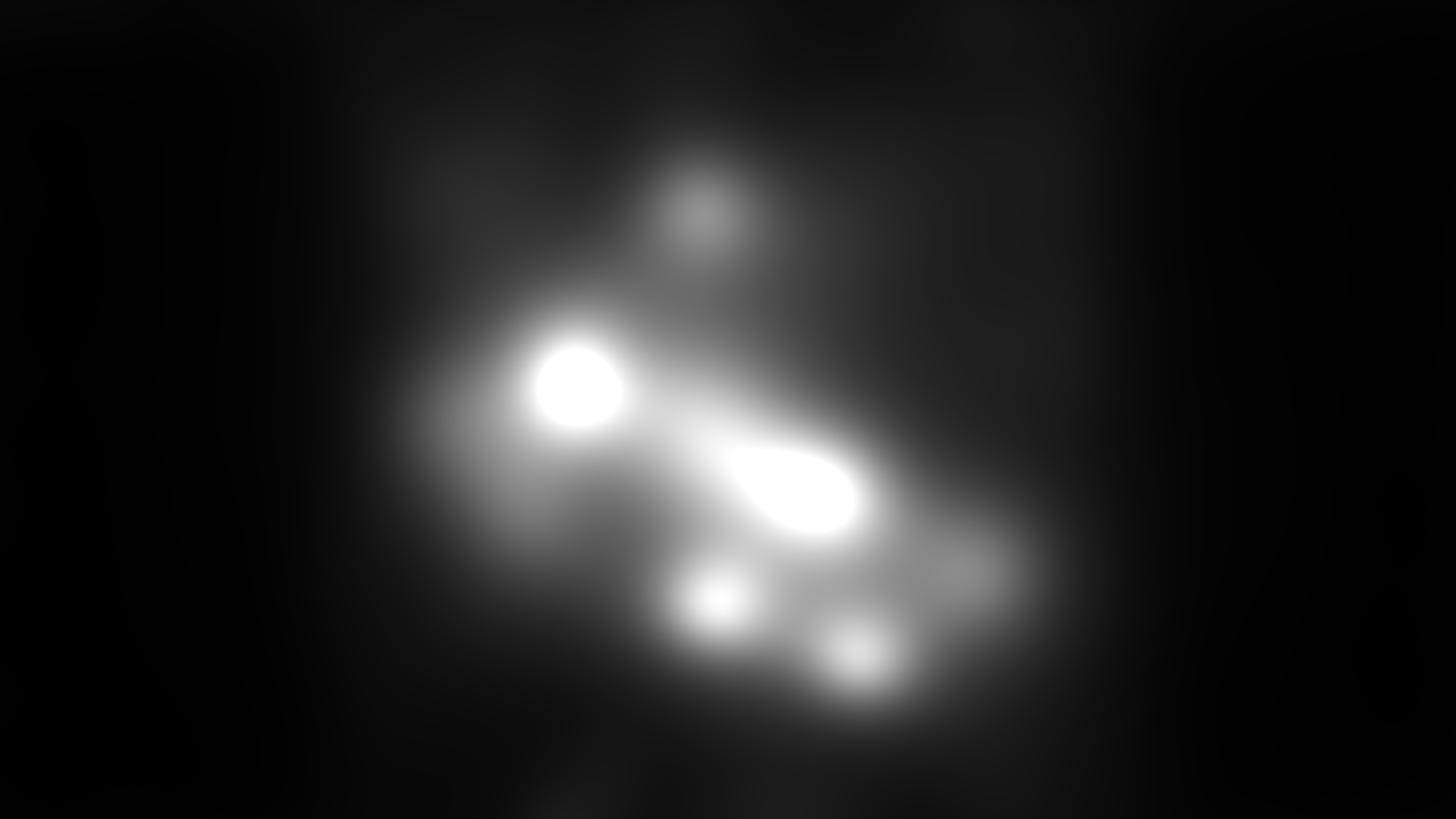} &
        \includegraphics[trim={9cm 0cm 9cm 0cm},clip,height=0.14\textwidth]{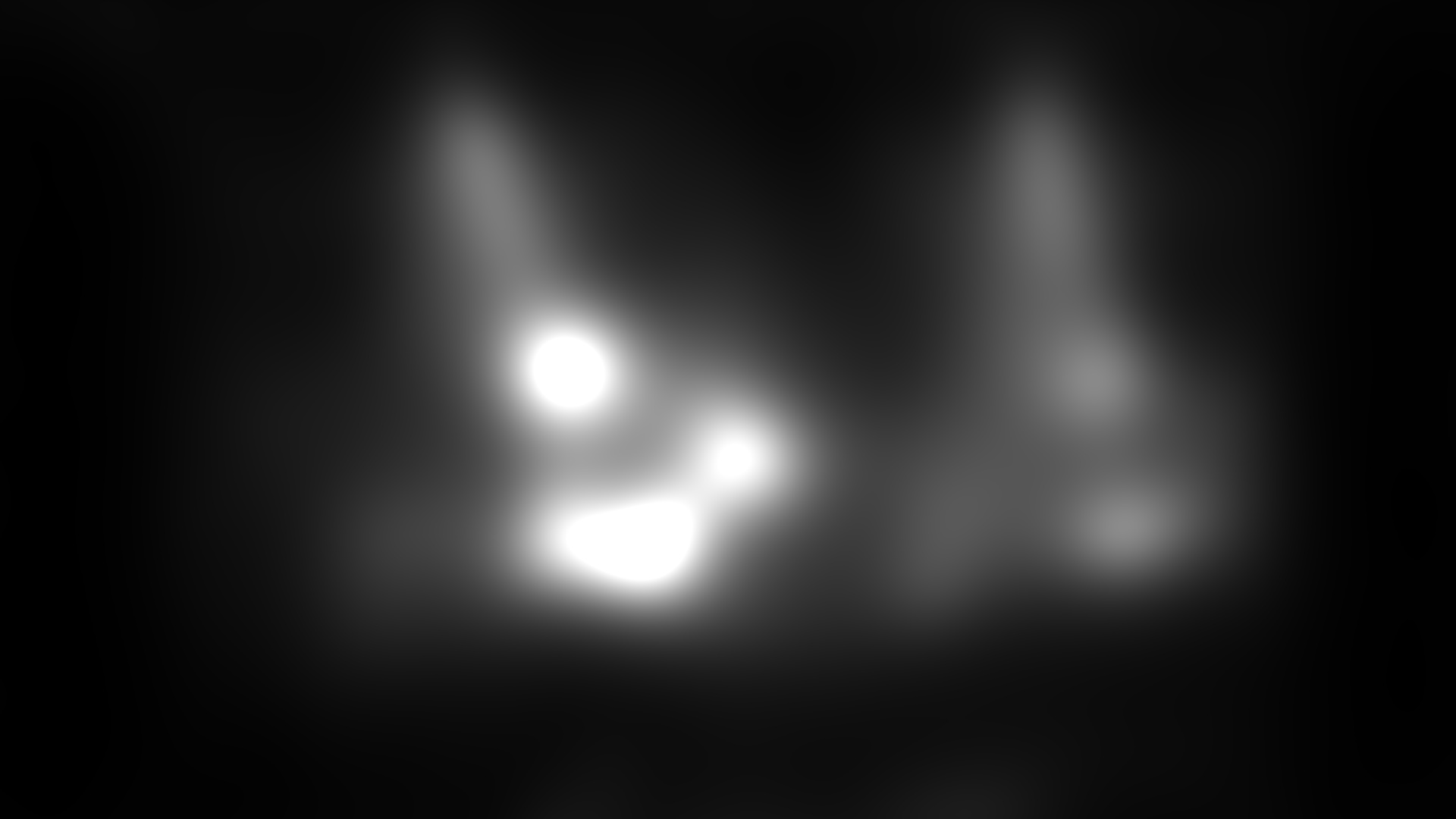} \\
        \rotatebox{90}{\textsf{\textbf{DenseGaze}}} \rotatebox{90}{\hspace{.35cm}(577ms)} &
        \includegraphics[trim={8cm 0cm 8cm 0cm},clip,height=0.14\textwidth]{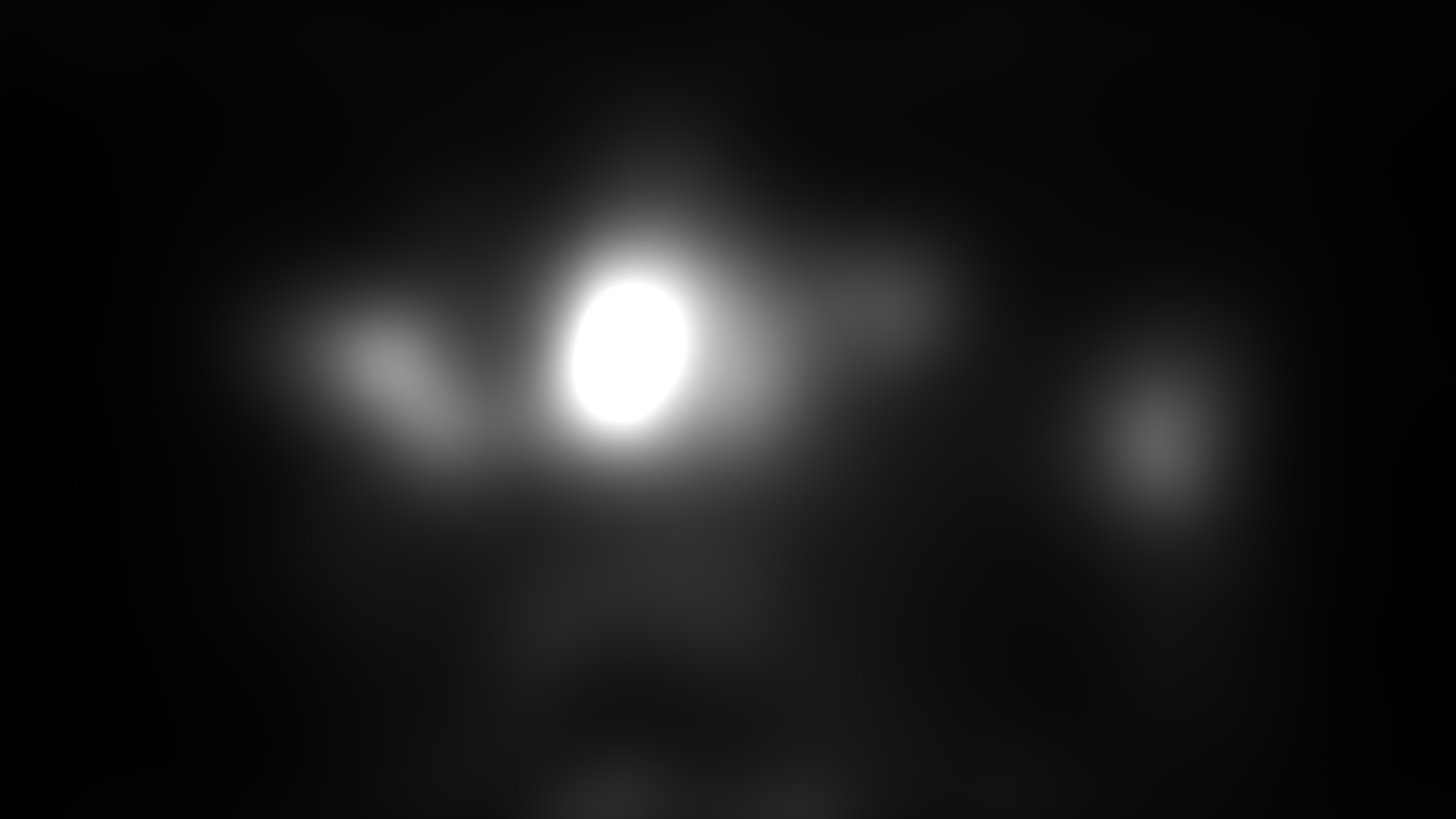} &
        \includegraphics[trim={3cm 0cm 3cm 0cm},clip,height=0.14\textwidth]{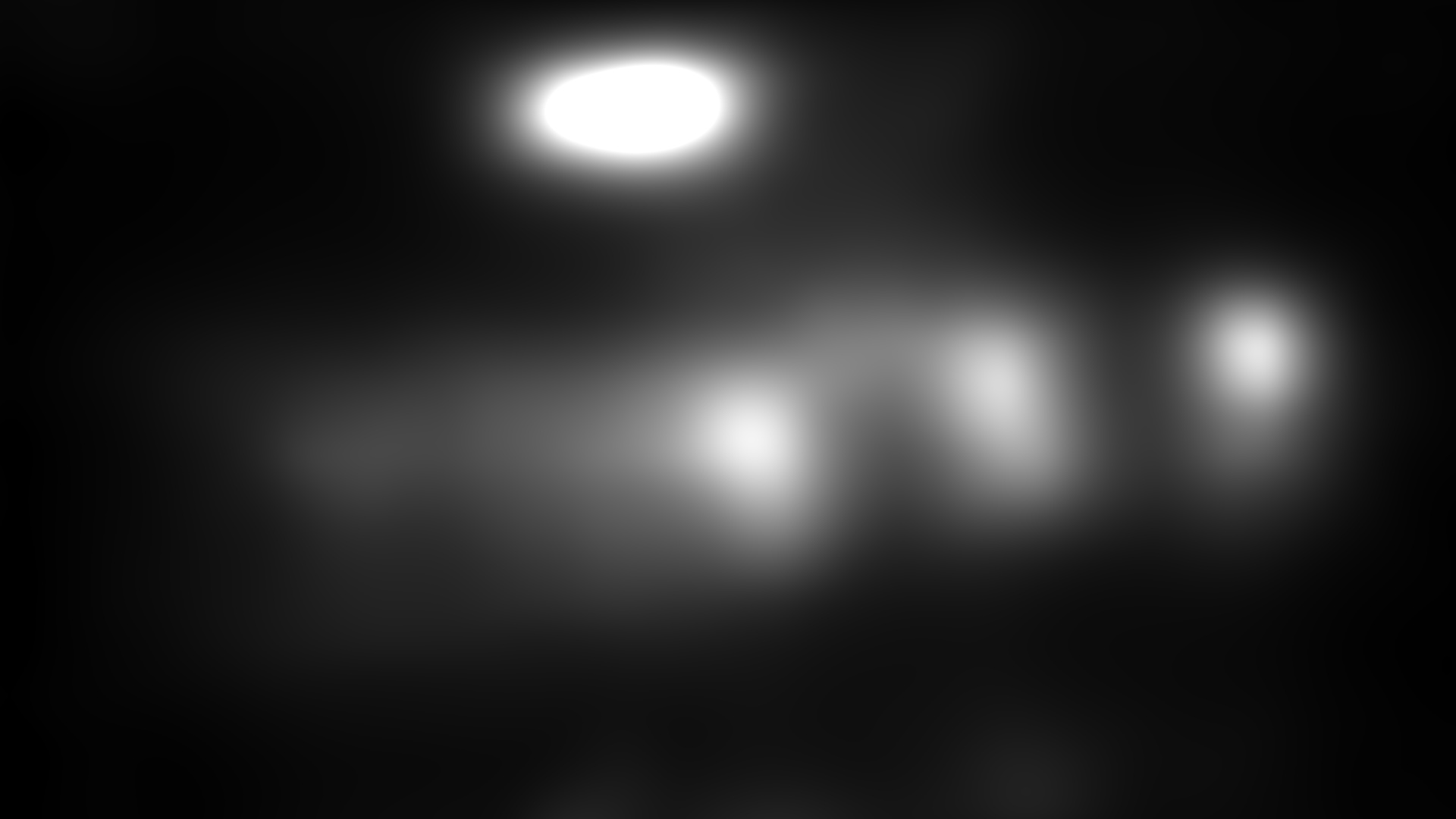} &
        \includegraphics[trim={9cm 0cm 9cm 0cm},clip,height=0.14\textwidth]{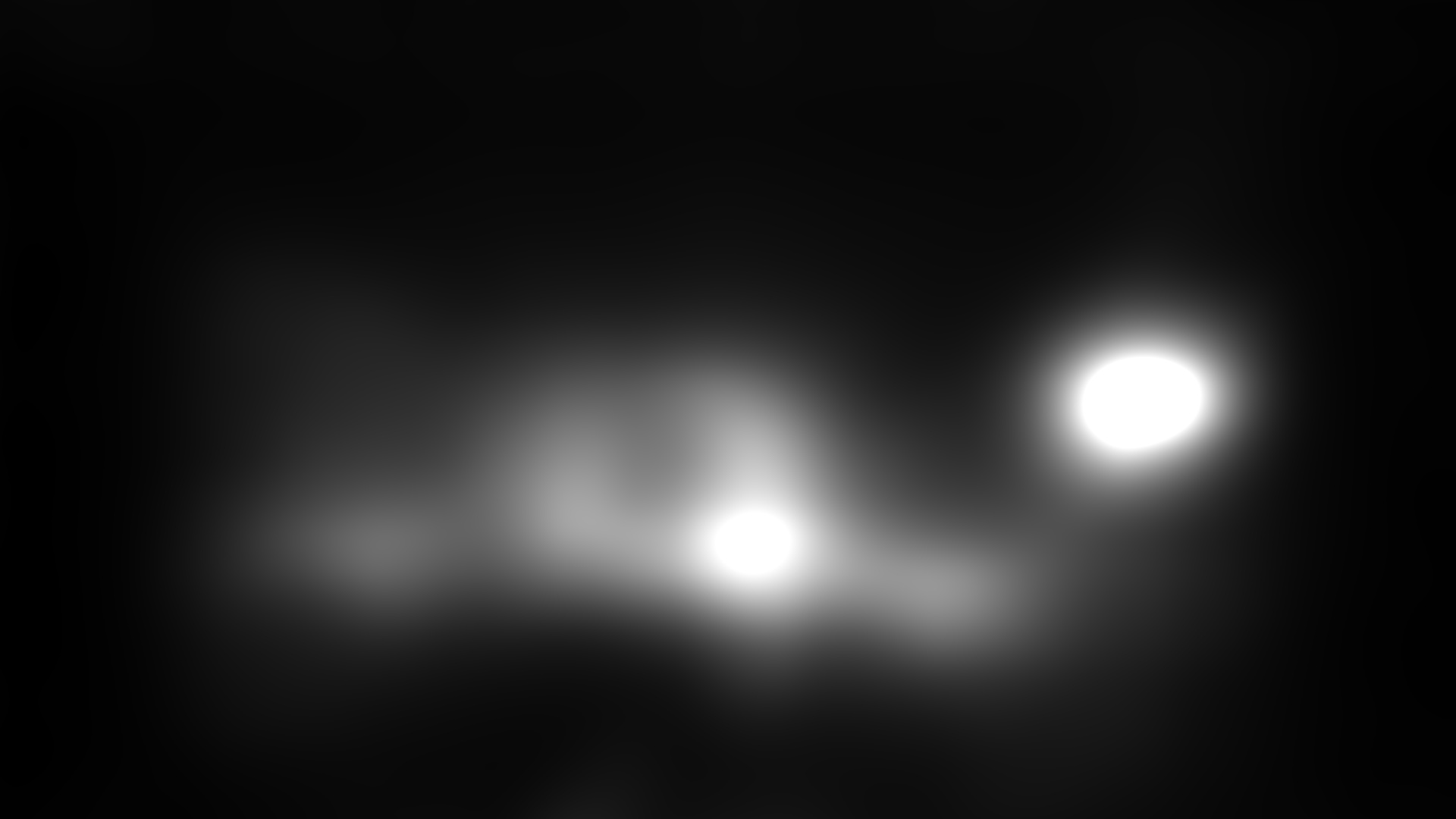} &
        \includegraphics[trim={18cm 0cm 18cm 0cm},clip,height=0.14\textwidth]{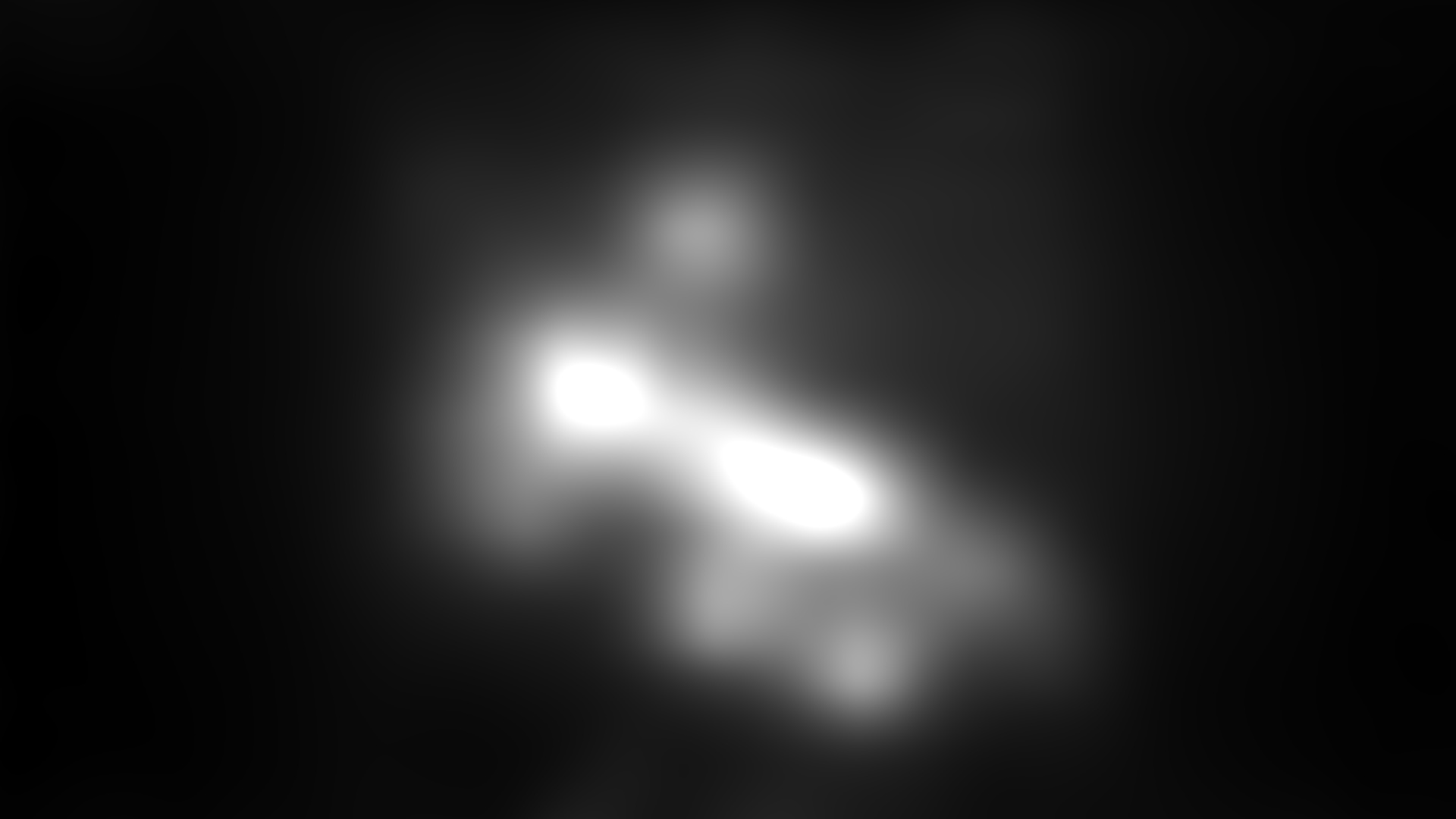} &
        \includegraphics[trim={9cm 0cm 9cm 0cm},clip,height=0.14\textwidth]{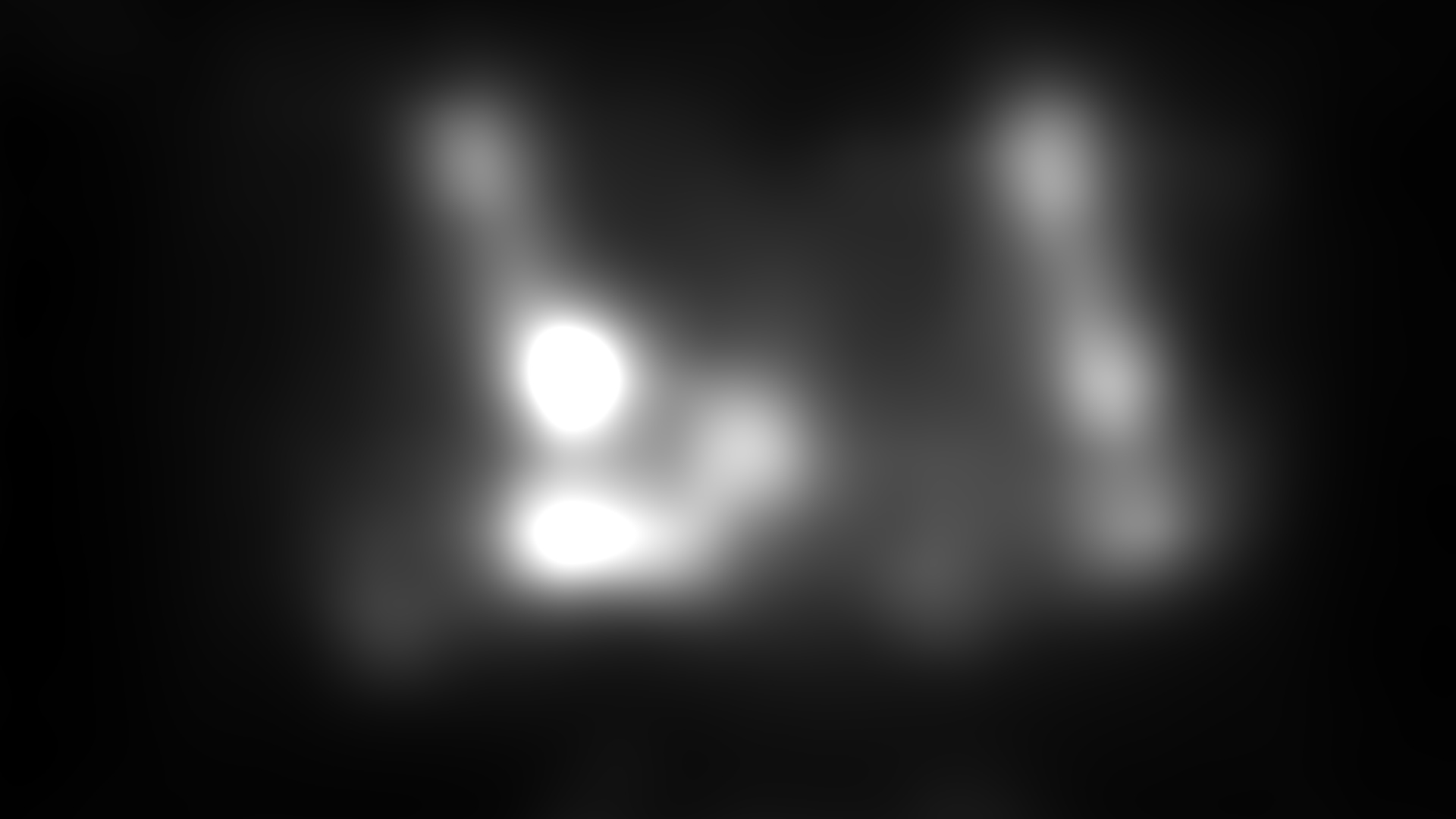} \\
        \rotatebox{90}{\hspace{0.15cm}\textsf{\textbf{FastGaze}}} \rotatebox{90}{\hspace{.35cm}(356ms)} &
        \includegraphics[trim={8cm 0cm 8cm 0cm},clip,height=0.14\textwidth]{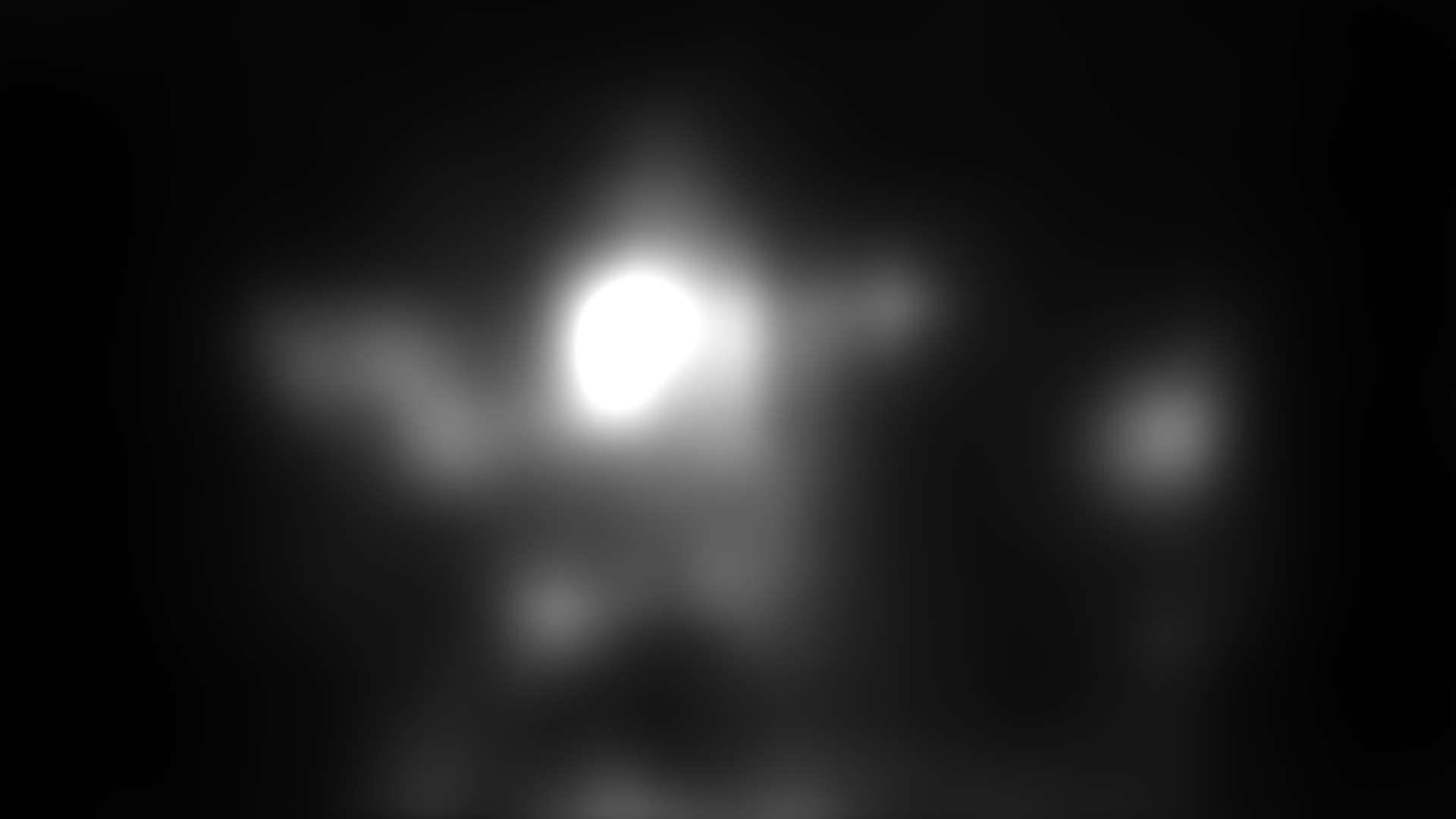} &
        \includegraphics[trim={3cm 0cm 3cm 0cm},clip,height=0.14\textwidth]{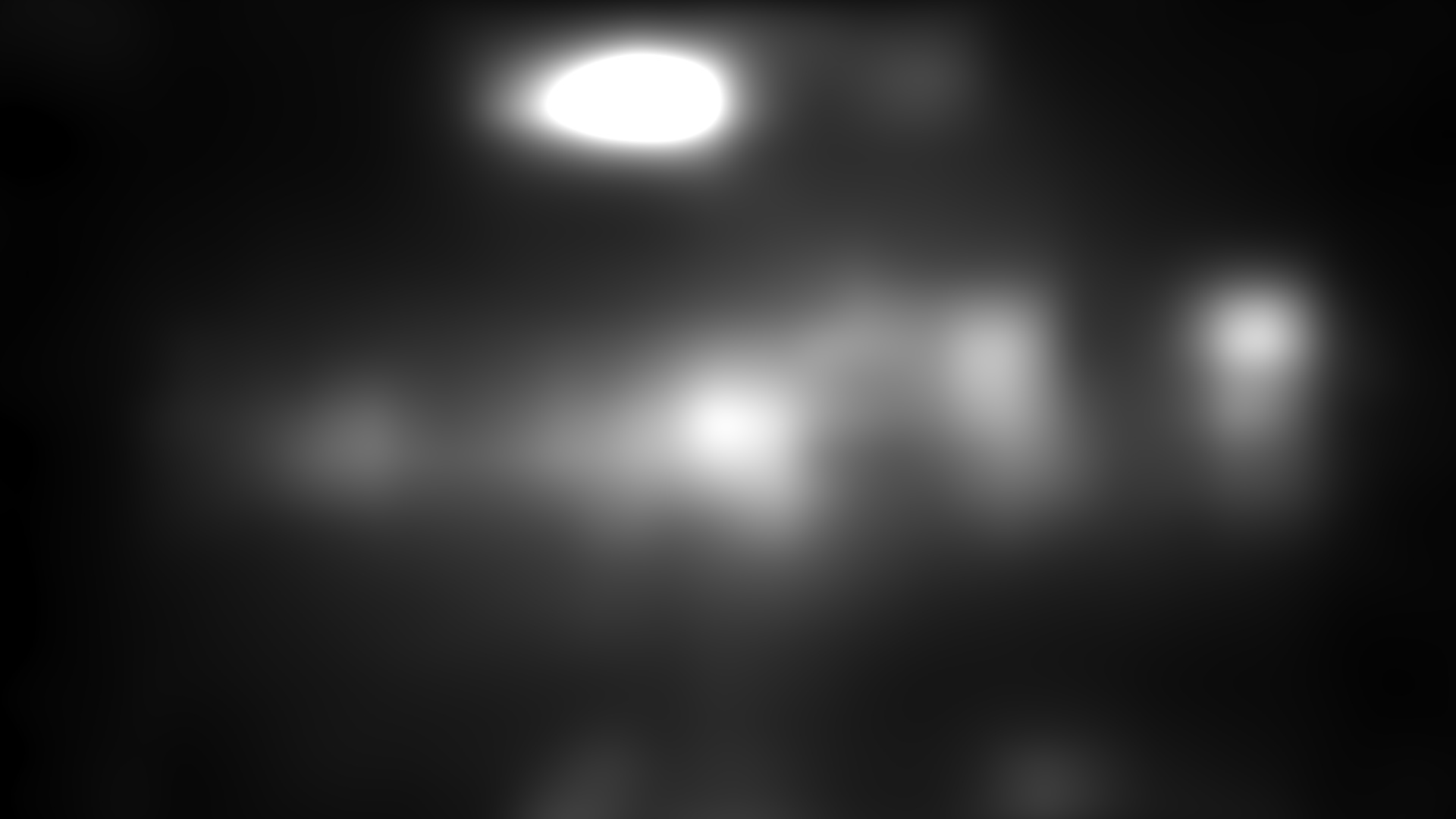} &
        \includegraphics[trim={9cm 0cm 9cm 0cm},clip,height=0.14\textwidth]{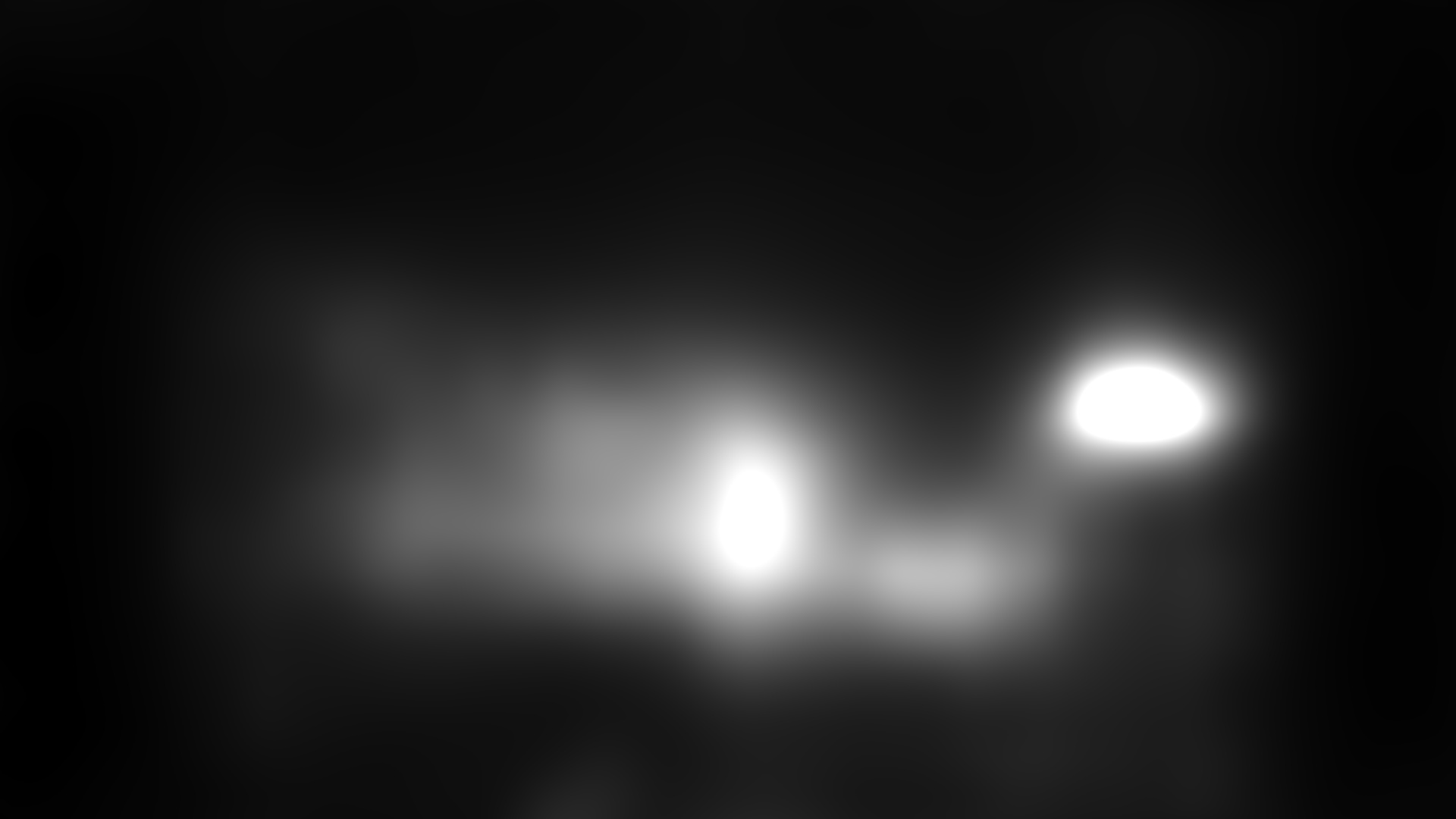} &
        \includegraphics[trim={18cm 0cm 18cm 0cm},clip,height=0.14\textwidth]{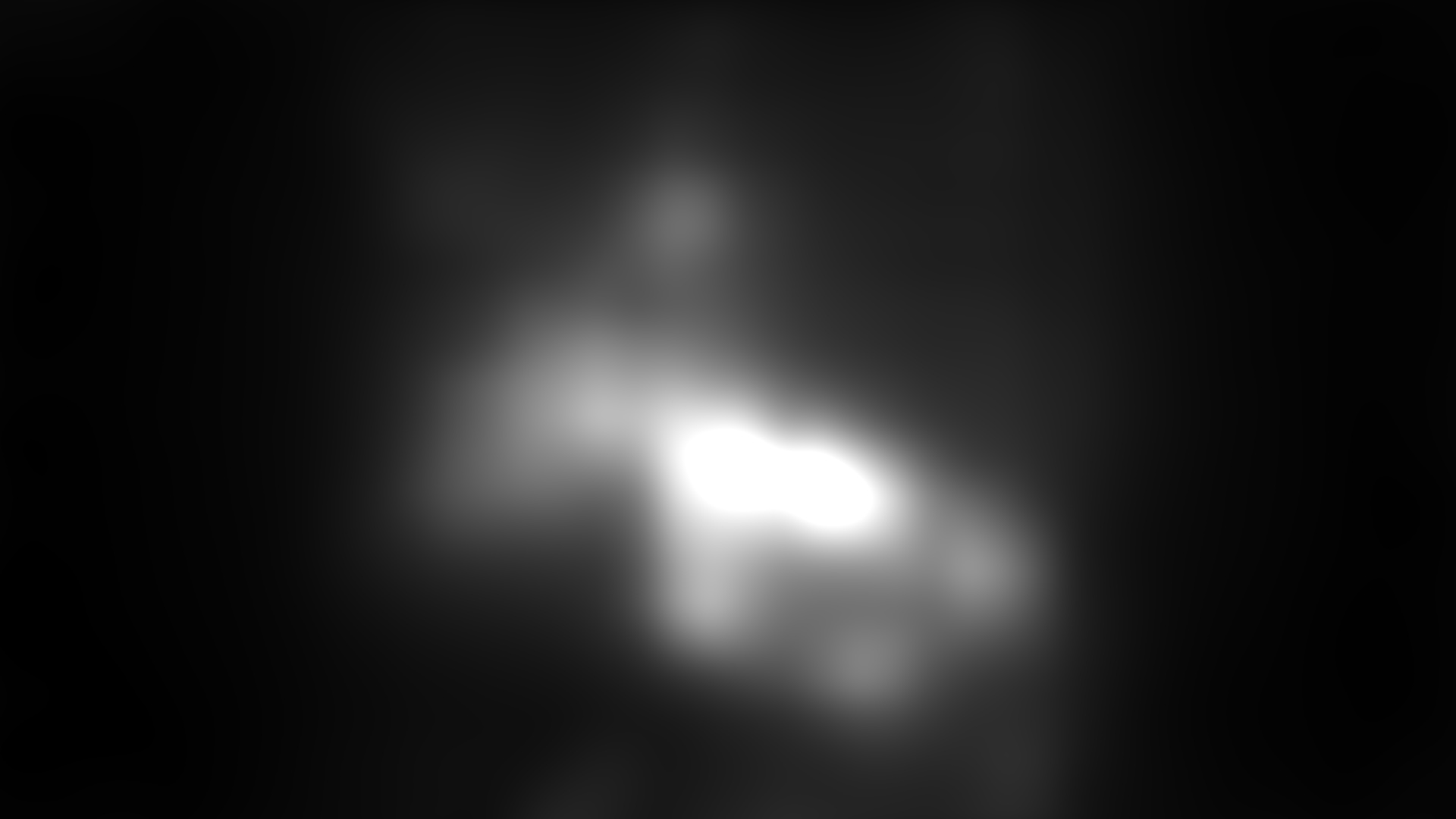} &
        \includegraphics[trim={9cm 0cm 9cm 0cm},clip,height=0.14\textwidth]{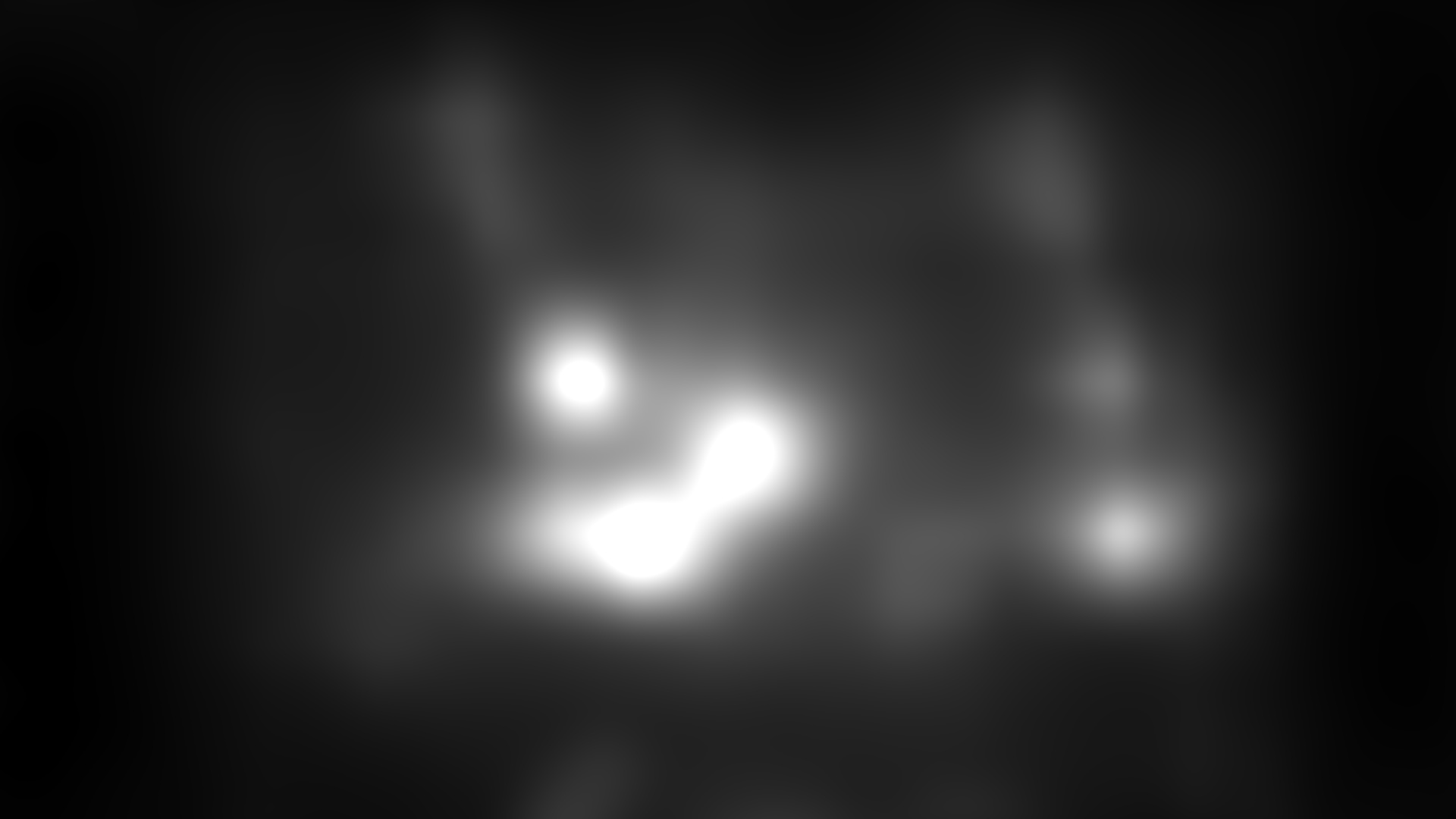} \\
        \rotatebox{90}{\hspace{0.15cm}\textsf{\textbf{FastGaze}}} \rotatebox{90}{\hspace{.4cm}(91ms)} &
        \includegraphics[trim={8cm 0cm 8cm 0cm},clip,height=0.14\textwidth]{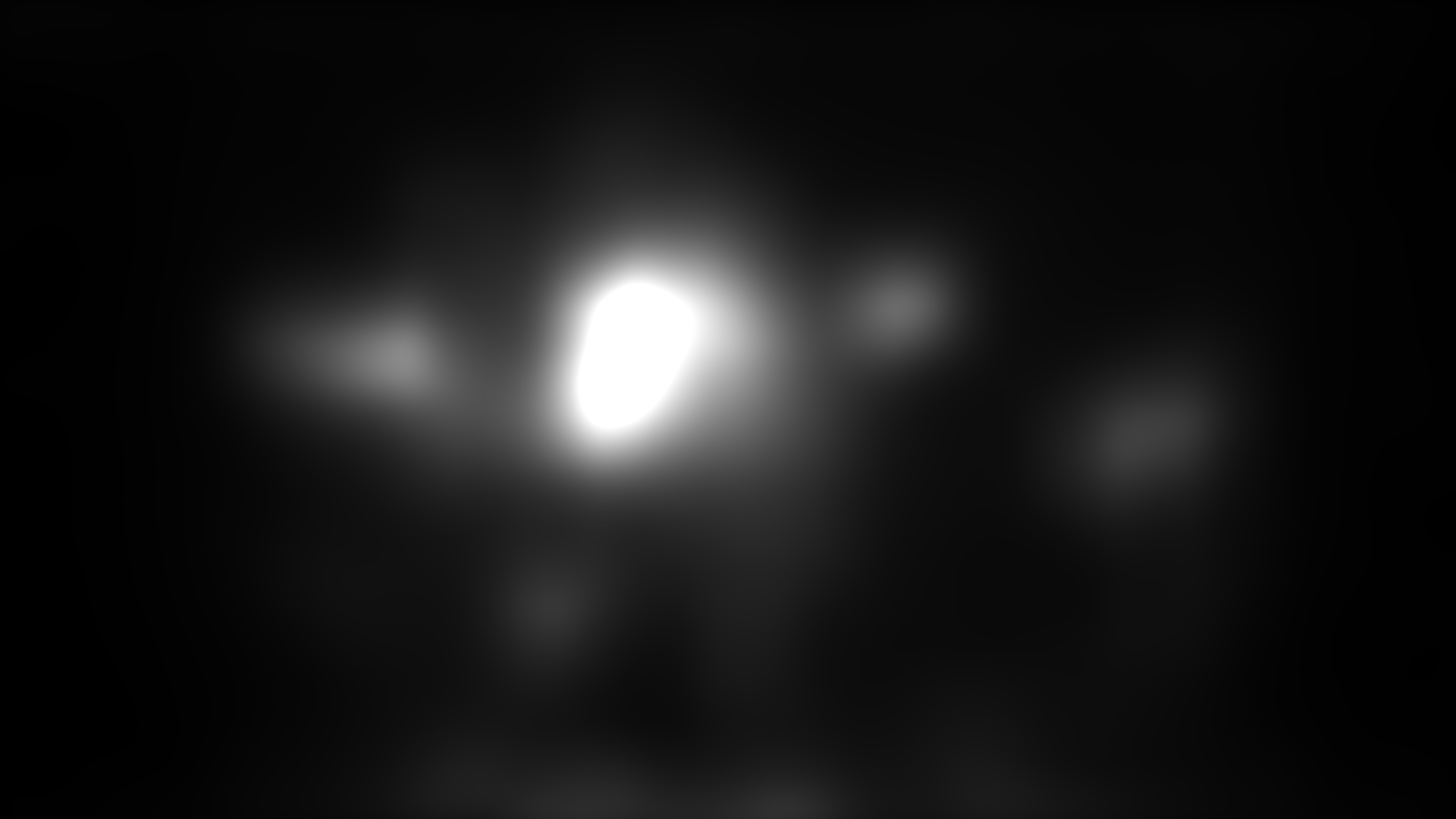} &
        \includegraphics[trim={3cm 0cm 3cm 0cm},clip,height=0.14\textwidth]{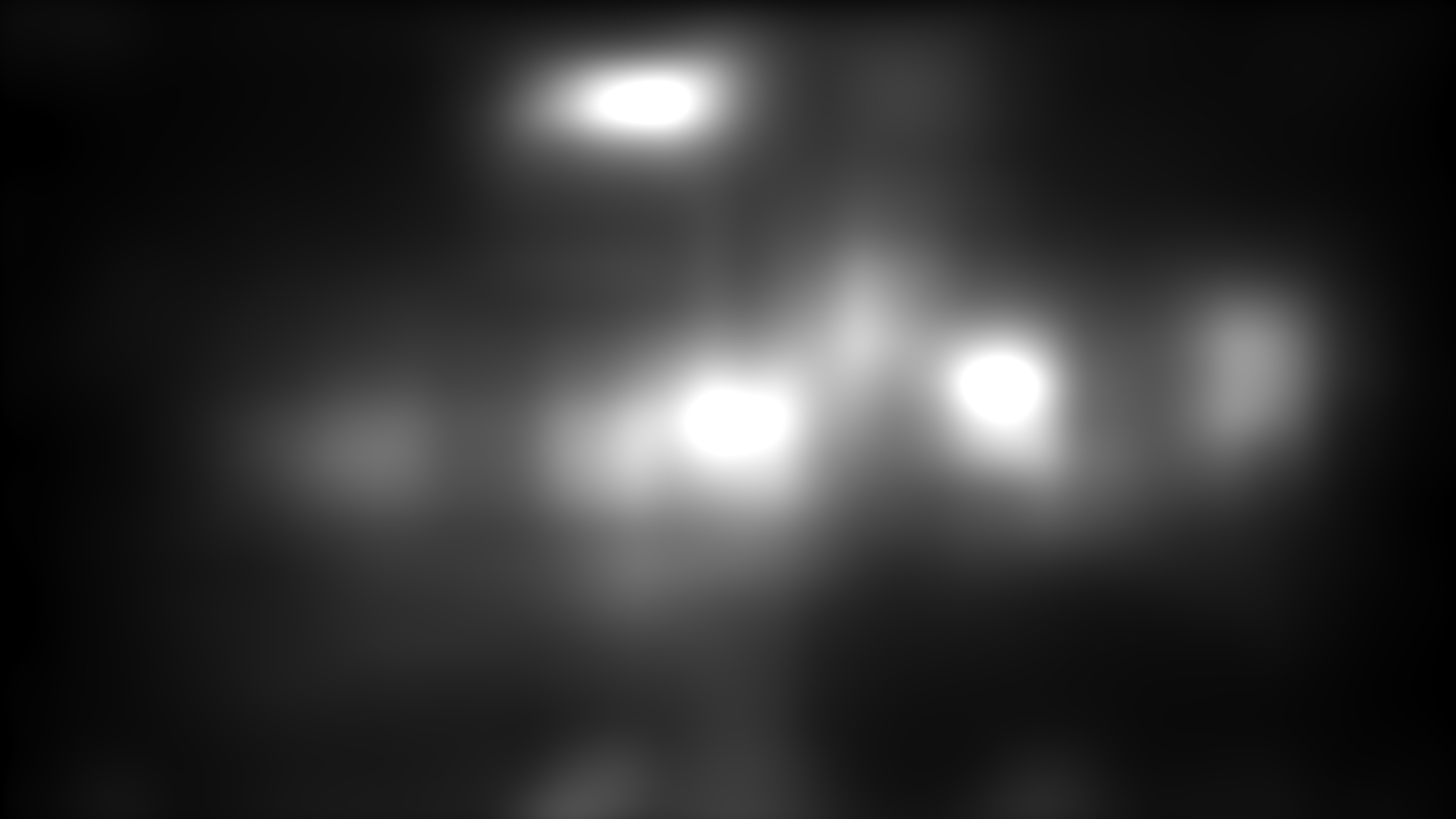} &
        \includegraphics[trim={9cm 0cm 9cm 0cm},clip,height=0.14\textwidth]{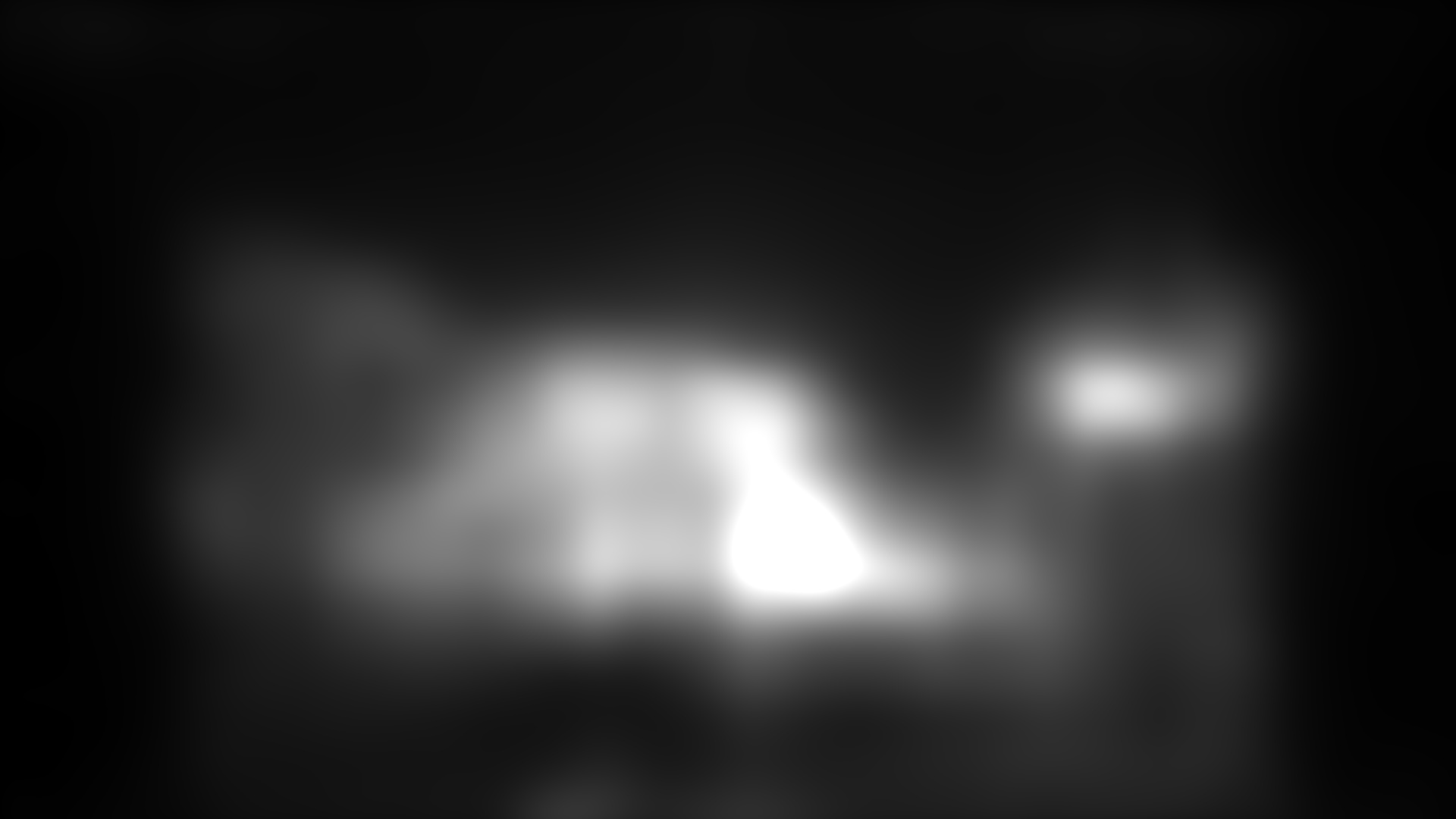} &
        \includegraphics[trim={18cm 0cm 18cm 0cm},clip,height=0.14\textwidth]{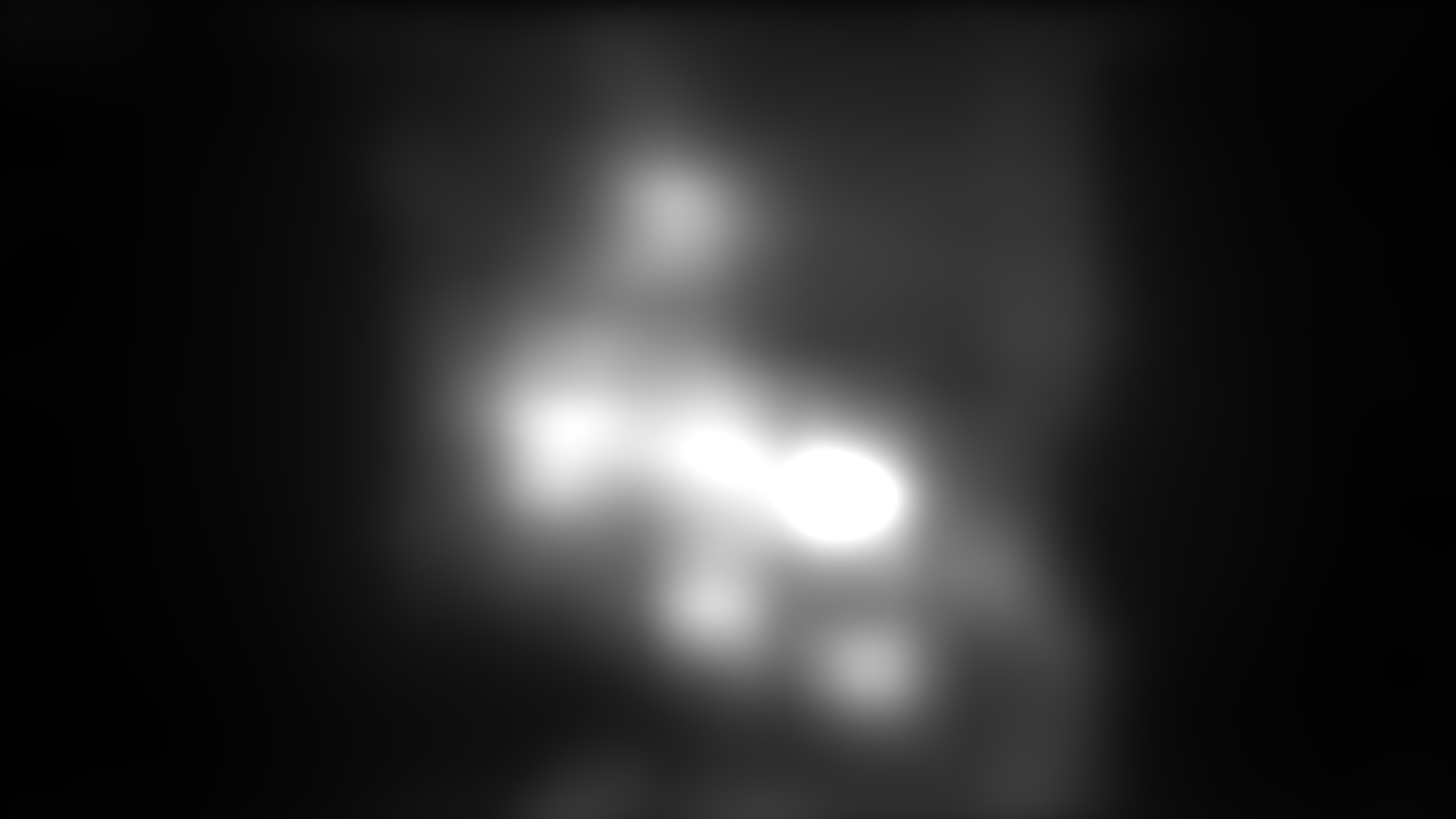} &
        \includegraphics[trim={9cm 0cm 9cm 0cm},clip,height=0.14\textwidth]{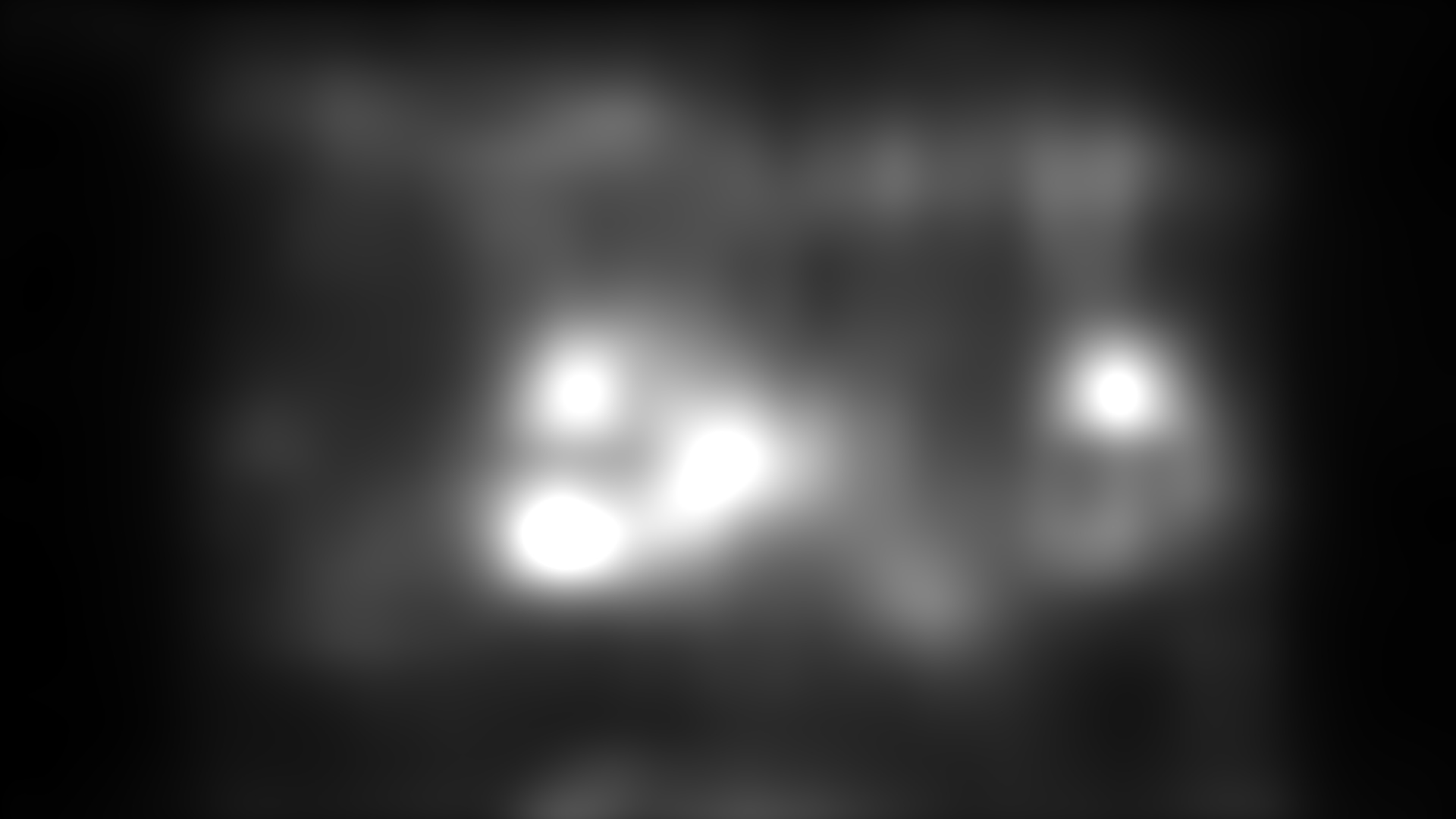}
      \end{tabular}
      \caption{Example saliency maps generated for images of the CAT2000 dataset. (The original
        images contained gray borders which were cropped in this visualization.) Intensities
        correspond to gamma-corrected fixation probabilities, where black and white correspond to the 1\% and
        99\% percentile of saliency values for a given image, respectively. Saliency maps in
        the second row were generated by an ensemble of DeepGaze~II models, but we report
        the time for a single model. Rows 3 and 4 show saliency maps of only slightly pruned DenseGaze
        and FastGaze models, while the last row shows saliency maps of a heavily pruned model. We
        find all models produce similar saliency maps. In particular, even the heavily pruned
        model (39x speedup compared to DeepGaze~II) still responds to faces, people, other objects,
        and text.}
      \label{fig:saliency_maps}
    \end{figure*}

    We find that explicitly regularizing computational complexity is important. For the same AUC
    and depending on the amount of pruning, we observe speedups of up to 2x for FastGaze
    when comparing regularized and non-regularized models.

    In Figure~\ref{fig:pruned_models} we visualize some of the pruned FastGaze models. We find that at
    lower computational complexity, optimal architectures have a tendency to alternate between
    convolutions with large and small numbers of feature maps. This makes sense when
    only considering the computational cost of a convolutional layer (Equation~\ref{eq:comp_cost}), but
    it is interesting to see that such an architecture can still perform well in terms of fixation
    prediction, which requires the detection of various types of objects.

    Qualitative results are provided in Figure~\ref{fig:saliency_maps}. Even at large reductions in
    computational complexity, the fixation predictions appear very similar. At a speedup of 39x
    compared to DeepGaze~II, the saliency maps start to become a bit blurrier, but generally detect
    the same structures. In particular, the model still responds to faces, people, objects, signs,
    and text.

    \begin{table}[t]
      \centering
      \caption{Comparison of deep models evaluated on the MIT300 benchmark dataset. For reference, we also include
        the performance of a Gaussian center bias. Results of competing methods were obtained from the
        benchmark's website \cite{bylinskii-mit}. The last column contains the computational cost in terms
        of the number of floating point operations required to process a $640 \times 480$ pixel input}
      \begin{tabular}{l|ccccc}
        \hline
        Model                                    & AUC           & KL            & SIM           & NSS           & GFLOP \\
        \hline
        Center Bias                              & 78\%          & 1.24          & 0.45          & 0.92          & - \\
        eDN \cite{vig2014edn}                    & 82\%          & 1.14          & 0.41          & 1.14          & - \\
        SalNet \cite{pan2016salnet}              & 83\%          & 0.81          & 0.52          & 1.51          & - \\
        DeepGaze~I \cite{kuemmerer2014deepgaze}  & 84\%          & 1.23          & 0.39          & 1.22          & - \\
        SAM-ResNet \cite{cornia2016sam}          & 87\%          & 1.27          & \textbf{0.68} & 2.34          & - \\
        DSCLRCN                                  & 87\%          & 0.95          & \textbf{0.68} & \textbf{2.35} & - \\
        DeepFix \cite{kruthiventi2017deepfix}    & 87\%          & 0.63          & 0.60          & 2.26          & - \\
        SALICON \cite{huang2015salicon}          & 87\%          & \textbf{0.54} & 0.60          & 2.12          & - \\
        DeepGaze~II \cite{kummerer2016deepgaze2} & \textbf{88}\% & 0.96          & 0.46          & 1.29          & 240.6 \\
        \textbf{FastGaze}                        & 85\%          & 1.21          & 0.61          & 2.00          & \textbf{10.7} \\
        \textbf{DenseGaze}                       & 86\%          & 1.20          & 0.63          & 2.16          & 12.8 \\
        \hline
      \end{tabular}
      \label{tbl:benchmark}
    \end{table}

    To verify that our models indeed perform close to the state of the art, we submitted saliency
    maps to the MIT Saliency Benchmark \cite{bylinskii-mit,judd2012benchmark,bylinskii2016}. We computed saliency maps for the
    MIT300 test set, which contains 300 more images of the same type as MIT1003.
    We evaluated a FastGaze model which took 356ms to evaluate in PyTorch, (2250 pruned
    features, $\beta = 0.0001$) and a DenseGaze model which took 577ms (2701 pruned features,
    $\beta = 3 \cdot 10^{-5}$).
    We find that both models perform slightly below the state of the art when evaluated on MIT300
    (Table~\ref{tbl:benchmark}), but are still comparable to other recent deep saliency models. We
    explain the discrepancy by the fact that the submitted models were chosen for their performance on
    CAT2000. That is, they generalize very well to other datasets, but may have lost information
    about the subtleties of the MIT datasets.

  \section{Conclusion}
   We have described a principled pruning method which only requires gradients as input, and which is
   efficient and easy to implement. Unlike most pruning methods, we explicitly penalized computational complexity and
   tried to find the architecture which optimally optimizes a given trade-off between performance
   and computational complexity. With this we were able to show that the computational complexity of
   state-of-the-art saliency models can be drastically reduced while maintaining a
   similar level of performance. Together with a knowledge distillation approach, the reduced
   complexity allowed us to train the models end-to-end and achieve good generalization performance.

   In settings where training is expensive, trying out many different parameters to tune the trade-off between
   computational complexity and performance may not be feasible.
   We have discussed an alternative pruning signal which takes into account computational complexity but is free
   of hyperparameters. This approach does not only apply to Fisher pruning, but can be combined with any pruning
   signal estimating the importance of a feature map or parameter.

   Less resource intensive models are of particular importance in applications where a lot of data
   is processed, as well as in applications running on resource constrained devices such as mobile
   phones. Faster gaze prediction models also have the potential to speed up the development of
   video models. The larger number of images to be processed in videos impacts training
   times, making it more difficult to iterate models. Another issue is that the amount of fixation
   training data in existence is fairly limited for videos. Smaller models will allow for faster training times and a more efficient
   use of the available training data.

  \clearpage

  \bibliographystyle{splncs}
  \bibliography{saliency}

\begin{thebibliography}{10}

\bibitem{ardizzone2013cropping}
Ardizzone, E., Bruno, A., Mazzola, G.
\newblock In: Saliency Based Image Cropping. Springer Berlin Heidelberg (2013)
  773--782

\bibitem{feng2012encoding}
Feng, Y., Cheung, G., Tan, W.T., Ji, Y.:
\newblock Gaze-driven video streaming with saliency-based dual-stream
  switching.
\newblock In: IEEE Visual Communications and Image Processing (VCIP). (2012)

\bibitem{xu2016gui}
Xu, P., Sugano, Y., Bulling, A.:
\newblock Spatio-temporal modeling and prediction of visual attention in
  graphical user interfaces.
\newblock In: Proceedings of the 2016 CHI Conference on Human Factors in
  Computing Systems. (2016)

\bibitem{koch1985neuro}
Koch, C., Ullman, S.:
\newblock Shifts in selective visual attention: towards the underlying neural
  circuitry.
\newblock Human Neurobiology \textbf{4} (1985)  219--227

\bibitem{kuemmerer2014deepgaze}
K{\"u}mmerer, M., Theis, L., Bethge, M.:
\newblock {Deep Gaze I: Boosting Saliency Prediction with Feature Maps Trained
  on ImageNet}.
\newblock In: ICLR Workshop. (May 2015)

\bibitem{kummerer2016deepgaze2}
{K{\"u}mmerer}, M., {Wallis}, T.S.A., {Bethge}, M.:
\newblock {DeepGaze II: Reading fixations from deep features trained on object
  recognition}.
\newblock ArXiv e-prints (October 2016)

\bibitem{hinton2015distilling}
{Hinton}, G., {Vinyals}, O., {Dean}, J.:
\newblock {Distilling the Knowledge in a Neural Network}.
\newblock ArXiv e-prints (2015)

\bibitem{molchanov2017pruning}
Molchanov, P., Tyree, S., Karras, T., Aila, T., Kautz, J.:
\newblock Pruning convolutional neural networks for resource efficient
  inference.
\newblock In: International Conference on Learning Representations (ICLR).
  (2017)

\bibitem{simonyan2014vgg}
{Simonyan}, K., {Zisserman}, A.:
\newblock {Very Deep Convolutional Networks for Large-Scale Image Recognition}.
\newblock ArXiv e-prints (September 2014)

\bibitem{nair2010relu}
Nair, V., Hinton, G.:
\newblock {Rectified linear units improve restricted boltzmann machines}.
\newblock In: Proceedings of the 27th International Conference on Machine
  Learning. (2010)

\bibitem{huang2017densely}
Huang, G., Liu, Z., van~der Maaten, L., Weinberger, K.Q.:
\newblock Densely connected convolutional networks.
\newblock In: The IEEE Conference on Computer Vision and Pattern Recognition
  (CVPR). (2017)

\bibitem{lecun1990optimal}
LeCun, Y., Denker, J.S., Solla, S.A.:
\newblock {Optimal Brain Damage}.
\newblock In Touretzky, D.S., ed.: Advances in Neural Information Processing
  Systems 2.
\newblock Morgan-Kaufmann (1990)  598--605

\bibitem{hassibi1993second}
Hassibi, B., Stork, D.G.:
\newblock {Second order derivatives for network pruning: Optimal brain
  surgeon}.
\newblock In: Advances in Neural Information Processing Systems. (1993)
  164--171

\bibitem{kingma2015adam}
Kingma, D.P., Ba, J.:
\newblock Adam: A method for stochastic optimization.
\newblock In: Proceedings of the 3rd International Conference on Learning
  Representations (ICLR). (2015)

\bibitem{jiang2015salicon}
Jiang, M., Huang, S., Duan, J., Zhao, Q.:
\newblock {SALICON: Saliency in Context}.
\newblock In: The IEEE Conference on Computer Vision and Pattern Recognition
  (CVPR). (June 2015)

\bibitem{judd2009mit}
Judd, T., Ehinger, K., Durand, F., Torralba, A.:
\newblock Learning to predict where humans look.
\newblock In: IEEE International Conference on Computer Vision (ICCV). (2009)

\bibitem{zhang2017rethinking}
Zhang, C., Bengio, S., Hardt, M., Recht, B., Vinyals, O.:
\newblock Understanding deep learning requires rethinking generalization.
\newblock In: International Conference on Learning Representations (ICLR).
  (2017)

\bibitem{kruthiventi2017deepfix}
Kruthiventi, S.S.S., Ayush, K., Babu, R.V.:
\newblock {DeepFix: A Fully Convolutional Neural Network for Predicting Human
  Eye Fixations}.
\newblock In: IEEE Transactions on Image Processing. Volume~26. (2017)

\bibitem{tavakoli2017}
Tavakoli, H.R., Borji, A., Laaksonen, J., Rahtu, E.:
\newblock Exploiting inter-image similarity and ensemble of extreme learners
  for fixation prediction using deep features.
\newblock Neurocomputing \textbf{244}(Supplement C) (2017)  10 -- 18

\bibitem{liu2016}
Liu, N., Han, J.:
\newblock A deep spatial contextual long-term recurrent convolutional network
  for saliency detection.
\newblock CoRR \textbf{abs/1610.01708} (2016)

\bibitem{vig2014edn}
Vig, E., Dorr, M., Cox, D.:
\newblock {Large-Scale Optimization of Hierarchical Features for Saliency
  Prediction in Natural Images}.
\newblock In: The IEEE Conference on Computer Vision and Pattern Recognition
  (CVPR). (2014)

\bibitem{pan2016salnet}
Pan, J., Sayrol, E., Giro-i Nieto, X., McGuinness, K., O'Connor, N.E.:
\newblock Shallow and deep convolutional networks for saliency prediction.
\newblock In: The IEEE Conference on Computer Vision and Pattern Recognition
  (CVPR). (June 2016)

\bibitem{li2017pruning}
Li, H., Kadav, A., Durdanovic, I., Samet, H., Graf, H.P.:
\newblock Pruning filters for efficient convnets.
\newblock In: International Conference on Learning Representations (ICLR).
  (2017)

\bibitem{han2015pruning}
Han, S., Pool, J., Tran, J., Dally, W.J.:
\newblock Learning both weights and connections for efficient neural networks.
\newblock In: Advances in Neural Information Processing Systems. (2015)

\bibitem{he2015prelu}
He, K., Zhang, X., Ren, S., , Sun, J.:
\newblock {Delving Deep into Rectifiers: Surpassing Human-Level Performance on
  ImageNet Classification}.
\newblock In: IEEE International Conference on Computer Vision (ICCV). (2015)

\bibitem{lecun1998lenet}
LeCun, Y., Bottou, L., Bengio, Y., Haffner, P.:
\newblock Gradient-based learning applied to document recognition.
\newblock In: Proceedings of the IEEE. Volume~86. (1998)  2278--2324

\bibitem{Kuemmerer2017a}
Kümmerer, M., Wallis, T.S.A., Bethge, M.:
\newblock Saliency benchmarking: Separating models, maps and metrics.
\newblock arxiv (Apr 2017)

\bibitem{borji2015cat2000}
Borji, A., Itti, L.:
\newblock {CAT2000: A Large Scale Fixation Dataset for Boosting Saliency
  Research}.
\newblock CVPR 2015 workshop on "Future of Datasets" (2015) arXiv preprint
  arXiv:1505.03581.

\bibitem{PyTorch}
PyTorch.
\newblock \url{https://github.com/pytorch}

\bibitem{bylinskii-mit}
Bylinskii, Z., Judd, T., Borji, A., Itti, L., Durand, F., Oliva, A., Torralba,
  A.:
\newblock {MIT Saliency Benchmark}

\bibitem{cornia2016sam}
Cornia, M., Baraldi, L., Serra, G., Cucchiara, R.:
\newblock Predicting human eye fixations via an lstm-based saliency attentive
  model.
\newblock CoRR \textbf{abs/1611.09571} (2016)

\bibitem{huang2015salicon}
Huang, X., Shen, C., Boix, X., Zhao, Q.:
\newblock {SALICON: Reducing the semantic gap in saliency prediction by
  adapting deep neural networks}.
\newblock In: The IEEE International Conference on Computer Vision (ICCV).
  (2015)

\bibitem{judd2012benchmark}
Judd, T., Durand, F., Torralba, A.:
\newblock A benchmark of computational models of saliency to predict human
  fixations.
\newblock Technical report, MIT technical report (2012)

\bibitem{bylinskii2016}
Bylinskii, Z., Judd, T., Oliva, A., Torralba, A., Durand, F.:
\newblock What do different evaluation metrics tell us about saliency models?
\newblock CoRR \textbf{abs/1604.03605} (2016)

\end{thebibliography}

  \newpage
  \renewcommand{\thesection}{S\arabic{section}}
  \setcounter{section}{0}
  \section{Details of Fisher pruning}

  \label{sec:pruning_details}
  Under mild regularity conditions, the diagonal of the Hessian of the cross-entropy loss is given by
  \begin{align}
    H_{kk}
    &= \frac{\partial^2}{\partial\theta_k^2} \mathbb{E}_P\left[ -\log Q_{\bm{\theta}}(\mathbf{z} \mid \mathbf{I}) \right] \\
    &= \frac{\partial}{\partial\theta_k} \mathbb{E}_P\left[ - \frac{1}{Q_{\bm{\theta}}(\mathbf{z} \mid \mathbf{I})} \frac{\partial}{\partial\theta_k} Q_{\bm{\theta}}(\mathbf{z} \mid \mathbf{I}) \right] \\
    &= \mathbb{E}_P\left[ \left( \frac{\partial}{\partial\theta_k} \log Q_{\bm{\theta}}(\mathbf{z} \mid \mathbf{I}) \right)^2 \right]
    - \mathbb{E}_P\left[ \frac{\frac{\partial^2}{\partial\theta_k^2} Q_{\bm{\theta}}(\mathbf{z} \mid \mathbf{I})}{Q_{\bm{\theta}}(\mathbf{z} \mid \mathbf{I})} \right],
  \end{align}
  where the last step follows from the quotient rule. For the second term we have
  \begin{align}
    \mathbb{E}_P\left[ \frac{\frac{\partial^2}{\partial^2\theta_k} Q_{\bm{\theta}}(\mathbf{z} \mid \mathbf{I})}{Q_{\bm{\theta}}(\mathbf{z} \mid \mathbf{I})} \right]
    &= \int P(\mathbf{I}) \frac{P(\mathbf{z} \mid \mathbf{I})}{Q_{\bm{\theta}}(\mathbf{z} \mid \mathbf{I})} \frac{\partial^2}{\partial\theta_k^2} Q_{\bm{\theta}}(\mathbf{z} \mid \mathbf{I}) \, d(\mathbf{I}, \mathbf{z}) \\
    &\approx \int P(\mathbf{I}) \frac{\partial^2}{\partial\theta_k^2} Q_{\bm{\theta}}(\mathbf{z} \mid \mathbf{I}) \, d(\mathbf{I}, \mathbf{z}) \\
    &= \frac{\partial^2}{\partial\theta_k^2} \int P(\mathbf{I}) Q_{\bm{\theta}}(\mathbf{z} \mid \mathbf{I}) \, d(\mathbf{I}, \mathbf{z}) \\
    &= 0
  \end{align}
  where we have assumed that $Q_{\theta}(\mathbf{z} \mid \mathbf{I})$ has been trained to convergence and is close to $P(\mathbf{z} \mid \mathbf{I})$.

  \section{Alternative derivation of Fisher pruning}
  \label{sec:alternative}

  Let $Q_{\theta}$ be our original model and $\tilde{Q}_{\theta,\mathbf{m}}$ be the pruned model,
  where we multiply the activations $a_{nkij}$ by binary mask parameters $m_k\in\{0,1\}$ as in
  Eqn.~8 of the main text. Pruning is achieved by setting $m_k = 0$ for pruned features, and $m_k=1$ for features we wish to keep.

  We can define the cost of pruning as the extent to which it changes the model's output, which can be measured by the KL divergence
  \begin{equation}
    \mathcal{L}(\mathbf{m}) = \frac{1}{N}\sum_{n=1}^N \operatorname{KL}[Q_\theta(\mathbf{z} \mid \mathbf{I}_n) \mid\mid \tilde{Q}_{\theta,\mathbf{m}}(\mathbf{z} \mid \mathbf{I}_n)].
  \end{equation}

  This KL divergence can be approximated locally by a quadratic distance (the Fisher-Rao distance) as we will show below. First, note that when $\mathbf{m}=\mathbf{1}$, $\tilde{Q}_{\theta,\mathbf{m}} \equiv Q_\theta$, so the value of the $\operatorname{KL}$ divergence is 0, and its gradients with respect to both $\theta$ and $\mathbf{m}$ are exactly $\mathbf{0}$ as well.

  Thus, we can approximate $\mathcal{L}$ by its second-order Taylor-approximation around the unpruned model $\mathbf{m}=\mathbf{1}$ as follows:
  \begin{equation}
    \mathcal{L}(\mathbf{m}) \approx \frac{1}{2} (\mathbf{m} - \mathbf{1})^\top \mathbf{H} (\mathbf{m} - \mathbf{1}),
  \end{equation}
  where $\mathbf{H} = \nabla^2 \mathcal{L}(\mathbf{1})$ is the Hessian of $\mathcal{L}$ at $\mathbf{m}=\mathbf{1}$.

  Pruning a single feature $k$ amounts to setting $\mathbf{m}_k = \mathbf{1} - \mathbf{e}_k$, where $\mathbf{e}_k$ is the unit vector which is zero everywhere except at its $i^{\text{th}}$ entry, where it is $1$. The cost of pruning a single feature $k$ is then approximated as:
  \begin{equation}
   \ell_k = \mathcal{L}(\mathbf{1} - \mathbf{e}_k) \approx \frac{1}{2}H_{kk}
  \end{equation}

  Under some mild conditions, the Hessian of $\mathcal{L}$ at $\mathbf{m}=\mathbf{1}$ is the Fisher information matrix, which can be approximated by the empirical Fisher information. In particular for the diagonal terms $H_{kk}$ we have that:
  \begin{align}
    \frac{1}{2}H_{kk} &= -\frac{\partial^2}{\partial m_k^2} \frac{1}{2N} \sum_{n=1}^N \mathbb{E}_{Q_{\theta,\mathbf{1}}} \left[  \log \tilde{Q}_{\theta,\mathbf{1}} (\mathbf{z} \mid \mathbf{I}_n) \right]\\
    &= \frac{1}{2N} \sum_{n=1}^N \mathbb{E}_{\tilde{Q}_{\theta,\mathbf{1}}} \left[\left( \frac{\partial}{\partial m_k} \log \tilde{Q}_{\theta,\mathbf{1}}(\mathbf{z} \mid \mathbf{I}_n) \right)^2 \right]\\
    &= \frac{1}{2N} \sum_{n=1}^N \mathbb{E}_{Q_{\theta}} \left[\left( a_{n,i,j,k} \frac{\partial}{\partial \tilde{a}_{n,i,j,k}} \log \tilde{Q}_{\theta,\mathbf{1}}(\mathbf{z} \mid \mathbf{I}_n) \right)^2 \right]\\
    &= \frac{1}{2N} \sum_{n=1}^N  a^2_{n,i,j,k} \mathbb{E}_{Q_{\theta}} \left[\left( \frac{\partial}{\partial a_{n,i,j,k}} \log Q_{\theta}(\mathbf{z} \mid \mathbf{I}_n) \right)^2 \right]\\
    &\approx \frac{1}{2N} \sum_{n=1}^N  a^2_{n,i,j,k} \left( \frac{\partial}{\partial a_{n,i,j,k}} \log Q_{\theta}(\mathbf{z_n} \mid \mathbf{I}_n) \right)^2\\
    &= \frac{1}{2N} \sum_{n=1}^N  g^2_{nk},
  \end{align}
  where $g_{nk}$ is defined as in Eqn.~9 of the main text.
\end{document}